\documentclass{article}

\usepackage{PRIMEarxiv}

\usepackage[utf8]{inputenc} 
\usepackage[T1]{fontenc}    
\usepackage{hyperref}       
\usepackage{url}            
\usepackage{booktabs}       
\usepackage{amsfonts}       
\usepackage{nicefrac}       
\usepackage{microtype}      
\usepackage{lipsum}
\usepackage{fancyhdr}       
\usepackage{graphicx}       
\graphicspath{{media/}}     

\usepackage{graphicx}
\usepackage{subfigure}
\usepackage{amsmath,amsthm,amssymb}
\usepackage{thmtools}
\usepackage{multirow}
\usepackage{bbm}
\usepackage{authblk}
\usepackage{kotex}
\usepackage{multicol}
\usepackage{wrapfig,lipsum}
\usepackage{algorithmic}
\usepackage[ruled,vlined]{algorithm2e}
\usepackage{caption}
\usepackage{subcaption}
\usepackage{hhline}
\usepackage{nicematrix}
\usepackage{float}
\newtheorem{theorem}{Theorem}[section]

\newtheorem{corollary}[theorem]{Corollary}
\newtheorem{definition}[theorem]{Definition}

\newtheorem{remark}[theorem]{Remark}

\title{Meta-ANOVA: Screening interactions for interpretable machine learning}

\author{%
Yougchan Choi$^{1*}$ \quad Seokhun Park$^{2*}$ \quad Chanmoo Park$^{2*}$ \quad Dongha Kim$^{3}$ \quad Yongdai Kim$^{2}$ \\
$^1$Toss bank \quad $^2$Seoul National University \quad $^3$SungShin Women's University\\
\texttt{\{pminer32\}@gmail.com}\\
\texttt{\{shrdid,chanmoo13,ydkim903\}@snu.ac.kr}\\
\texttt{dongha0718@sungshin.ac.kr}
}

\begin{document}
\maketitle

\begin{abstract}
There are two things to be considered when we evaluate predictive models. One is prediction accuracy, and the other is interpretability. 
Over the recent decades, many prediction models of high performance, such as ensemble-based models and deep neural networks, have been developed. 
However, these models are often too complex, making it difficult to intuitively interpret their predictions. 
This complexity in interpretation limits their use in many real-world fields that require accountability, such as medicine, finance, and college admissions. 
In this study, we develop a novel method called Meta-ANOVA to provide an interpretable model for any given prediction model. 
The basic idea of Meta-ANOVA is to transform a given black-box prediction model to the functional ANOVA model.
A novel technical contribution of Meta-ANOVA is a procedure of
screening out unnecessary interactions before transforming a given black-box model to the functional ANOVA model. 
This screening procedure allows the inclusion of higher order interactions in the transformed functional ANOVA model without computational difficulties. We prove that the screening procedure is asymptotically consistent. Through various experiments with synthetic and real-world datasets, we empirically demonstrate the superiority of Meta-ANOVA.
\end{abstract}


\section{Introduction}

Machine learning models have received great attention due to their remarkable prediction accuracy in various fields, and the emergence of Deep Neural Networks (DNNs) has further accelerated this interest (\cite{he2016deep, radford2019language, shen2017deep, devlin2018bert, chouiekh2018convnets, sharif2014cnn, webb2018deep, beck2020overview}).
Despite their strong prediction power, their applications to real world problems are limited
due to the difficulty in interpreting the decision process of machine learning models.
Typically the improvement of prediction powers has been achieved through increased model complexities that make the interpretation harder.
Most popularly used machine learning models including DNNs are considered as "black-box" models because understanding how and why they make their final decisions is almost impossible.
While black-box models could be acceptable to low-risk tasks, they pose significant challenges in high-risk applications, such as cancer diagnosis and self-driving car systems, where interpretability is crucial. 
Due to the need of trustworthiness in modern machine learning models for real world applications, eXplainable Artificial Intelligence (XAI) has become an important research topic.

In this paper, we develop an algorithm called Meta-ANOVA which transfers a given black-box machine learning model to an interpretable model. 
Meta-ANOVA learns a functional ANOVA model (\cite{gu2013smoothing}) that approximates
a given black-box prediction model closely. 
The functional ANOVA model, which decomposes a high-dimensional function into the sum of low-dimensional interpretable functions so-called interactions, is considered as one of the most important XAI tools (\cite{caru2019purifying}). 
For a given black-box model $f$, Meta-ANOVA approximates $f$ by the sum of interactions:

\begin{equation}
\label{eq:MNOVA}
f(\bold{x}) \approx \beta_0+ \sum_{j=1}^p f_j(x_j)+ \sum_{j<k} f_{jk}(x_j,x_k) + \cdots.
\end{equation}

There are various algorithms to learn a functional ANOVA model from given training data (\cite{chong1993ssanova, kim2009boosting, lin2006component}).
A unique and novel feature of Meta-ANOVA is to learn a functional ANOVA model from a pre-trained black-box model instead of training data. An important advantage of using a pre-trained black-box model is to be able to screen unnecessary interactions before learning a functional ANOVA model.
When the dimension of the input features is large, the number of interactions in the functional ANOVA model becomes too large, making learning all interactions simultaneously 
computationally prohibitive.
To overcome this challenge, we develop a novel interaction screening algorithm by use of a pre-trained black-box model. 
The proposed screening algorithm can delete unnecessary interactions before learning the functional ANOVA model. 
For linear regression models, several feature screening methods (\cite{fan2008sure, tibshirani1996regression, wang2020simple}) have been proposed, but there is no existing method 
for screening interactions in the functional ANOVA model. 
Meta-ANOVA does interaction screening successively by utilizing the information in a given black-box model. 

The main contributions of this work are summarized as follows:
\begin{itemize}
    \item We propose an algorithm so-called Meta-ANOVA to approximate a given black-box machine learning model by
      the functional ANOVA model.
    \item We develop a novel interaction screening algorithm based on a given black-box machine learning model and an algorithm to learn the functional ANOVA model only with selected interactions.
    \item Theoretically, we prove the selection consistency of the proposed interaction screening algorithm.
    \item By analyzing simulated and real datasets, we illustrate that Meta-ANOVA is a useful
    tool for XAI.
\end{itemize}

This paper is organized as follows. 
In Section \ref{sec:related}, we briefly review related works.
In Section \ref{sec:method}, we provide the proposed method.
Results of various numerical experiments for Meta-ANOVA are presented in Section \ref{sec:experiment}, and conclusions follow in Section \ref{sec:conclusion}.

\section{Related works}
\label{sec:related}

In general, there is a trade-off between prediction accuracy and interpretability (\cite{linardatos2020explainable}).
Linear regression models and decision trees are interpretable machine learning models, but their prediction accuracies are limited. 
Modern machine learning models, including  ensembles and DNNs, have shown remarkable performance in prediction, but their interpretability is poor.
 
Interpretable machine learning methods can be roughly classified into  two groups. One is {\it transparent-box design} and the other is {\it post-hoc interpretation}. 

\begin{remark}
Explainability is a synonym of interpretability. 
There have been various attempts to distinguish these two terms (\cite{lipton2018mythos, doshi2017towards, gilpin2018explaining, montavon2018methods}).
However, despite these attempts, their definitions lack mathematical formality and rigorousness. 
Conceptually, interpretability is mostly connected with the intuition behind the outputs of a model. 
On the other hand, explainability is associated with the internal logic (\cite{linardatos2021explainable}). 
From these views, our method has both sides, and thus we use them interchangeably.
\end{remark}

\subsection{Transparent-box design}

Transparent-box design aims at learning a machine learning model that can be interpretable (so-called "white-box model"). 
White-box models simultaneously predict and interpret, which makes them reliable for their applications to real world problems.
However, in order to make a white-box model, constraints on the model should be imposed, which leads to performance degradation.
Most of recent studies for transparent-box design focus on DNNs.
Self-Explaining Neural Network (SENN, \cite{melis2018towards}) tries to learn self-explaining neural networks that satisfy some interpretable properties linear models have. 
\cite{zhang2018interpretable} aims to train filters in a high convolutional layer to represent objects so that the filter itself can be interpreted.
Prototypical network (ProtoPNET, \cite{li2018deep}) presents a new architecture for Convolutional Neural Network (CNN) to provide explanations for each prediction. 
Attention Branch Network (ABN, \cite{fukui2019attention}) introduces a branch structure with an attention mechanism to provide visualized explanations.
Self-Interpretable model with Transformation Equivariant Interpretation (SITE, \cite{wang2021self}) learns a self-interpretable model that produces explanations invariant to transformation.

Using neural networks to learn the functional ANOVA model has also received much attention.
Neural Additive Model (NAM, \cite{agarwal2020neural}) is a specially designed neural network for learning the Generalized Additive Model (GAM, \cite{hastie2017generalized}).
Moreover, Neural Basis Model (NBM, \cite{radenovic2022neural}) and NODE-GAM (\cite{chang2021node}) learn functional ANOVA model by considering interactions.

\subsection{Post-hoc interpretation}

Post-hoc interpretation is methods to try to understand the inference process of a given black-box model.
There is no performance degradation but interpretation is less reliable, and thus
interpretation reliability is a key issue for these methods (\cite{linardatos2020explainable}).

Post-hoc interpretation can be divided into model-specific and model-agnostic methods. 
Model-specific methods are designed to interpret specific model classes. 
Most of the methods focus on DNNs. \cite{zeiler2014visualizing} utilizes de-convolution (transposed convolution) to visualize intermediate convolution filters. \cite{simonyan2013deep} develops a method that computes the gradient of the class score of CNNs with respect to the input to calculate the saliency map, and since then many gradient based interpretation methods have been proposed (\cite{sundararajan2017axiomatic, shrikumar2017learning}).  
Class Activation Maps (CAM, \cite{zhou2016learning}) utilizes the global average pooling to indicate the discriminative regions in the input space and variants of CAM are proposed by \cite{selvaraju2017grad, chattopadhay2018grad, wang2020score}. 

Neural Interaction Detection (NID, \cite{tsang2017detecting}) statistically searches for interactions in a given neural network by examining the weight connection  between input and output.
Persistent Interaction Detection (PID, \cite{liu2020detecting}) re-defines the weight connection utilizing persistent homology theory to select interactions. 

Model-agnostic methods are designed to interpret any black-box models.
Local Interpretable Model-agnostic Explanations (LIME, \cite{ribeiro2016should}) and
its modification Deterministic LIME (\cite{zafar2019dlime}) interpret a given black-box model by locally approximating it by an interpretable linear model. 

Several methods to measure the importance of each input feature based on the Shapley value (\cite{shapley1953value}) have been proposed. 
SHAP (\cite{lundberg2017unified}) is an unified measure of feature importance based on the Shapley value, Bivariate Shapley (\cite{masoomi2021explanations}) captures important interactions using a directed graph, and Faith-Shap (\cite{tsai2023faith}) extends the Shapley value to include feature interactions up to a given maximum order.  
\cite{lundstrom2023unifying} compares the gradient based and Shapley value based methods 
by use of the newly defined synergy function.

\section{Proposed method}
\label{sec:method}

In this section, we present the Meta-ANOVA algorithm to interpret a given black-box machine learning model through a functional ANOVA model. 
A novelty of Meta-ANOVA is the ability to screen out unnecessary higher order interactions before training the functional ANOVA model and thus to incorporate higher order interactions.
In contrast, existing learning algorithms for the functional ANOVA model such as smoothing spline (\cite{gu2013smoothing}), NAM (\cite{agarwal2020neural}) and NBM (\cite{radenovic2022neural}) only include the main effects and/or second-order interactions into the model due to computational burden for considering higher order interactions, in particular, when the dimension of input features is large.
Interaction screening is essential for incorporating  higher order interactions into the model and Meta-ANOVA does this successfully.

In Section \ref{sec:importance}, we introduce a measure for importance of interactions called the importance score, and we propose a consistent estimator of the importance score in Section \ref{sec:estimation} and Section \ref{sec:imp_score_main}. 
In Section \ref{sec:learning_anova}, we implement the Meta-ANOVA algorithm with the estimated importance scores.
For technical simplicity, we only consider continuous input features. 
For binary input features, we can modify the Meta-ANOVA algorithm easily by replacing the partial derivative operator with the partial difference operator.
See Section \ref{app: cate} of Appendix for details.

\subsection{Importance score for interactions}
\label{sec:importance}

We introduce a measure for the importance of interactions.
Let $\bold{x} \in \mathcal{X}\subset \mathbb{R}^p$ be an input feature vector on the input space $\mathcal{X}$, where $\mathcal{X}=\prod_{j=1}^p \mathcal{X}_j$ and each $\mathcal{X}_j$ is a subset of $\mathbb{R}$.
Let $f_{0}:\mathcal{X} \rightarrow \mathbb{R}$ be the true model and  $f:\mathcal{X} \rightarrow \mathbb{R}$ be a given black-box model that estimates $f_0$.
We will refer to $f$ as "the baseline model" or "the baseline black-box model" in the remainder of this paper.
For technical simplicity, we assume $\mathcal{X}_j=[0,1].$ 
Let $[p] = \{1, \cdots, p\},$ and let $J_k = \{\bold{j}\subset[p]:|\bold{j}|=k\}$.
For a given index set $\bold{j}\subset [p]$ and $\bold{x}\in \mathcal{X}$, let $\bold{x}_{\bold{j}}=(x_j, j\in \bold{j})$ be the subvector of $\bold{x}$.
For the true function $f_0$, we consider the following functional ANOVA decomposition:
$$f_0(\bold{x}) = \beta_{0} + \sum_{k=1}^{p} \sum_{\bold{j} \in J_{k}} f_{0\bold{j}}(\bold{x}_{\bold{j}}).$$
For $\bold{j}\in J_k,$ we refer to $f_{0\bold{j}}$ as the $k^{\text{th}}$-order interaction
with respect to $\bold{x}_{\bold{j}}$.
Note that there are $p \choose k$ many $k^{\text{th}}$-order interactions in $f_0$, which becomes large when $p$ and $k$ are large.
Thus, estimating all possible interactions would be computationally prohibitive, and thus  screening interactions is indispensable to include higher order interactions.

For given $j \in [p]$, the partial derivative operator $D_j$ of $f_0$ at $\bold{x}$ with respect to the index $j$ is defined as:
\begin{align}
\label{eq:con}
D_{j}f_0(\bold{x}):= \lim_{\epsilon \rightarrow 0} \frac{f_0(\bold{x} + \epsilon \bold{e}_{j} )- f_0(\bold{x})}{\epsilon}
\end{align}
where $\bold{e}_{j}$ is the $p$-dimensional vector whose $j^{\text{th}}$ entry is 1 and the other entries are 0. 
For $\bold{j}=\{j_1,\ldots,j_k\} \subset [p]$, the partial derivative of $f_0$ at $\bold{x}$ with respect to $\bold{j}$ is defined as $D_{\bold{j}}f_0(\bold{x}):=D_{j_{1}}\circ \cdots \circ D_{j_{k}} f_0(\bold{x})$.
We assume that $D_{\bold{j}} f_0$ exists for all $\bold{j}\subset [p]$.

Let $\bold{X}$ be a random vector where the input feature vector $\bold{x}$ is considered as a realization of $\bold{X}$.
Let $P_{\bold{X}}$ be the joint distribution of $\bold{X}$ and let $P_{\bold{j}}$ be the joint distribution of $\bold{X}_{\bold{j}}$.
We write $\bold{j}' > \bold{j}$ when $\bold{j}' \supsetneq \bold{j}$.
The following theorem, which is the key result for this paper, gives a necessary and sufficient condition for certain unnecessary interactions being 0 simultaneously.
The proof is given in Section \ref{proof:conti} of Appendix.
\begin{theorem}
\label{thm:conti}
For given $\bold{j} \subset [p] $,
$f_{\bold{j}'} (\cdot) \equiv 0$ for all $\bold{j}' > \bold{j}$ if and only if \\
$$I(\bold{j}) := \mathbb{E}_{\bold{X}_{\bold{j}}' \sim \mathbb{P}_{\bold{j}} }\left[\operatorname{Var}_{\bold{X}_{\bold{j}^c}' \sim \mathbb{P}_{\bold{j}^c}}\left\{ D_{\bold{j}}f_{0}(\bold{X}_{\bold{j}}',\bold{X}_{\bold{j}^c}')|\bold{X}_{\bold{j}}' \right\}\right]=0.$$
\end{theorem}
Theorem \ref{thm:conti} suggests that we can remove unnecessary interactions based on the values of $I(\bold{j})$.
That is, we can remove all higher order interactions that include $\bold{j}$ when $I(\bold{j})=0$ .
In this sense, we refer to $I(\bold{j})$ as the importance score for the interactions higher than $\bold{j}$.
Of course, the importance score $I(\bold{j})$ is not observable since neither $f_0$ nor
$P$ are observable.
In the following subsection, we propose a consistent estimator of $I(\bold{j})$.

An important computational advantage of using $I(\bold{j})$ (or its estimate) for screening interactions is that we can remove all of the higher order interactions simultaneously (i.e., all $\bold{j}' > \bold{j}$).
In other words, $f_{0\bold{j}}$ is nonzero only when $I(\bold{j}')>0$ for all $\bold{j}' <\bold{j}$ .
This property makes it possible to develop an algorithm for screening interactions similar to the Apriori algorithm for association analysis (\cite{srikant1996fast}). 
\begin{remark}
  $I(\bold{j})$ is not an importance measure for $f_{0\bold{j}}$.
  Instead, it is an importance measure for $\{f_{0\bold{j}'}: \bold{j}'>\bold{j}\}.$
  Thus, even when $I(\bold{j})>0,$ there are unnecessary interactions among 
  $\{f_{0\bold{j}'}: \bold{j}'>\bold{j}\}.$ If we want to select signal interactions only, we need a post-processing procedure after interaction screening by $I(\bold{j}).$
  See Section \ref{app: score interaction} of Appendix for such a procedure.
\end{remark}

\subsection{Estimation of the importance scores}
\label{sec:estimation}

Let $\left\{(\bold{x}^{i}, y^{i})\right\}_{i=1}^{n}$ be the training data used for learning a given black-box model $f$, which are assumed to be independent realization of $(\bold{X},Y) \sim \mathbb{P}$.
In this subsection, we propose an estimator of $I(\bold{j})$ and prove that it is consistent under regularity conditions.
A technically difficult part in estimating $I(\bold{j})$ is to estimate $D_{\bold{j}}f_0(\cdot)$ from the baseline black-box model.

For $j\in [p],$ we propose to estimate $D_j f_0(\bold{x})$ by the following estimator:
\begin{align}
\label{deriv_est}
\widehat{D}_jf_{0}(\bold{x}) = \frac{f(\bold{x}+\frac{1}{2}h_1 \bold{e}_{j}) - f(\bold{x}-\frac{1}{2}h_1\bold{e}_{j})}{h_1}
\end{align}
where $h_1>0,$ the bandwidth parameter, is a hyper-parameter given by a user or selected by validation data.
The estimator (\ref{deriv_est}) is a modification of the local linear estimator introduced in multiple articles (\cite{stone1977consistent, tsybakov1986robust, fan1997local}).
While the original local linear estimator is defined on the training data $\left\{(\bold{x}^{i}, y^{i})\right\}_{i=1}^{n},$ our estimator uses the outputs of the black-box model $f$ instead of $y_i$s. 
That is, our estimator is the local linear estimator based on $\left\{(\bold{x}^{i}, f(\bold{x}^{i}))\right\}_{i=1}^{n}.$

We first prove the consistency of the proposed estimator (\ref{deriv_est}) of the first derivative and then extend it to higher order derivatives.
For this purpose, we assume that the true function $f_0$ is $p+1$ times differentiable and the sup-norms of its derivatives are bounded above by a constant $L>0$.
In addition, we assume that the baseline black-box model $f$ has its $L^{\infty}$-risk upper bounded by  $O(\psi_n)$, that is
\begin{align*}
    &\limsup _{n \rightarrow \infty}  \mathbb{E}_{f}\left[\psi_n^{-1}\left\|f-f_0\right\|_{\infty}^2\right] \leq C_0<\infty
\end{align*}
where $\|f\|_{\infty}=\sup_{\bold{x}\in\mathcal{X}} |f(\bold{x})|$ and $\mathbb{E}_{f}$ 
is the expectation operator with respect to the distribution of the training data.
\begin{remark}
It is known that the minimax rate of $\psi_n$ is $n^{-2\beta/(2\beta+p)}$ when $f_0$ belongs to the H\"{o}lder space with smoothness $\beta$, and an estimator based on DNNs with the ReLU activation function nearly achieves it \cite{imaizumi2023sup}.
\end{remark}
The following theorem gives the convergence rate of the estimator (\ref{deriv_est}). 
The proof is given in Section \ref{proof:deriv_est} of Appendix.
{\theorem{
\label{thm:deriv_est}
If the bandwidth $h_1$ is set to be $h_1=h_{1,n=}\alpha \psi_n^{1/4}$ for any $\alpha>0,$ we have
\begin{align*}
    \limsup _{n \rightarrow \infty}  \mathbb{E}_{f}\left[(\psi_n^*)^{-1}\left\|\widehat{D}_jf_{0}-D_jf_{0}\right\|_{\infty}^2\right] \leq C
\end{align*}
where $\psi^*_n = \psi_n^{1/2}$ and $C<\infty$ is a constant depending only on $p$ and $L$.
}}
Theorem \ref{thm:deriv_est} states that the rate $\psi_n^*$ is slower than $\psi_n$ by a factor of $\psi_n^{1/2}$. 
The slower convergence rate is unavoidable since the original local linear estimator does (\cite{fan1997local}). 
The convergence rate $\psi_n^*$ is by no means optimal.
We could improve this rate by use of a higher order local polynomial estimator instead of the local linear estimator. 
However, a higher order local polynomial estimator is computational demanding due to the matrix inversions.

We estimate a higher order partial derivative $D_{\bold{j}}f_0(\bold{x}) := D_{j_1} \circ \cdots \circ D_{j_k} f_0(\bold{x})$ by applying the estimator (\ref{deriv_est}) successively.
For example, we estimate $D_{\{j_1,j_2\}} f_0(\bold{x})$ by
\begin{align}
\label{deriv_est2}
\widehat{D}_{\{j_1,j_2\}} f_{0}(\bold{x}) = \frac{ \widehat{D}_{j_1} f_0(\bold{x}+\frac{1}{2}h_2\bold{e}_{j_2}) - 
\widehat{D}_{j_1} f_0 (\bold{x}-\frac{1}{2}h_2\bold{e}_{j_2})}{h_2}.
\end{align}
The next corollary gives the convergence rate of $\widehat{D}_{\bold{j}} f_0(\cdot),$ which is getting slower as the order of interaction increases. 
The proof can be done by applying Theorem \ref{thm:deriv_est} repeatedly.
{\corollary{
\label{cor:deriv_est}
Suppose we set the bandwidth $h_{k,n}$ for estimation of the $k$th-order partial derivatives
as $h_{k,n}=\psi_n^{1/2^{k+1}}$ for $k=1,\ldots,p.$
Then, we have
\begin{align}
   \label{eq:con_rate}
    \limsup _{n \rightarrow \infty}  \mathbb{E}_{f}\left[(\psi_{n,\bold{j}}^*)^{-1}\left\|\widehat{D}_{\bold{j}}f_{0}-D_{\bold{j}} f_{0}\right\|^2_{\infty}\right] \leq C
\end{align}
where $\psi^*_{n,\bold{j}} = \psi_n^{1/2^{|\bold{j}|}}$ and $C<\infty$ is a constant depending only on $p$  and $L$.
}}
Finally, we estimate $I(\bold{j})$ by $\hat{I}(\bold{j}),$ where
$$\hat{I}(\bold{j}) = \mathbb{E}_{\bold{X}_{\bold{j}}' \sim \hat{\mathbb{P}}_{\bold{j}} }\left[\operatorname{Var}_{\bold{X}_{\bold{j}^c}' \sim \hat{\mathbb{P}}_{\bold{j}^c}}\left\{ \widehat{D}_{\bold{j}}f_{0}(\bold{X}_{\bold{j}}',\bold{X}_{\bold{j}^c}')|\bold{X}_{\bold{j}}' \right\}\right]$$
and $\hat{\mathbb{P}}$ is the empirical distribution of $\mathbb{P}$. 
The convergence rate of $\hat{I}(\bold{j})$ to $I(\bold{j})$ is given in the following theorem, whose proof is in Section \ref{proof:imp_est} of Appendix.
{\theorem{
Suppose that $\widehat{D}_{\bold{j}} f_0$ satisfies (\ref{eq:con_rate}). Then
\label{thm:imp_est}
\begin{align*}
    \limsup _{n \rightarrow \infty}  \mathbb{E}_{f}\left[(\psi_{n,\bold{j}}^*)^{-1/2}\left|
     \hat{I}(\bold{j})-I(\bold{j}) \right|^2\right] \leq C',
\end{align*}
where $C'<\infty$ is a constant depending only on $C$, $p$  and $L$.  }}

\subsection{Importance scores for each input feature}
\label{sec:imp_score_main}

Note that  the importance scores introduced in the previous subsections are not applicable to screening unnecessary main effects since $I(j)=0$ does not imply $f_{0j}(\cdot) \equiv 0$.
However, a modification of the importance score in Theorem \ref{thm:conti} can be made for this purpose. 

Suppose that $f_0$ does not depend on feature $x_j$ at all.
Then, we have $\operatorname{Var}_{X_j\sim \mathbb{P}_j} (f_0(X_j, \bold{x}_{j^c}))=0$
for all $\bold{x}_{j^c},$ where $\bold{x}_{j^c}=\bold{x}_{[p]\backslash \{j\}}$.
The converse of this statement is also true. 
That is, if $\operatorname{Var}_{X_j\sim \mathbb{P}_j} (f_0(X_j, \bold{x}_{j^c}))=0$
for all $\bold{x}_{j^c},$ then $f(\bold{x})$ does not depend on $x_j$.
So we can delete all the main effects and interactions involving $j$ when
$I^{(0)} (j) := \mathbb{E}_{\bold{X}_{j^c}\sim \mathbb{P}_{j^c}}\left[\operatorname{Var}_{X_j\sim \mathbb{P}_j} (f_0(X_j, \bold{X}_{j^c})|\bold{X}_{j^c})\right]=0$.
This quantity is the same as the total effect considered by \cite{saltelli2010variance}.
To estimate $I^{(0)} (j)$, we replace $f_0$ and $\mathbb{P}$ by the given black-box model $f$ and the empirical distribution, respectively.
Consistency can be proved similarly to that of $\hat{I}(j)$ and thus we omit it.

\subsection{Meta-ANOVA algorithm}
\label{sec:learning_anova}

The Meta-ANOVA algorithm consists of two steps.
In the first step, unnecessary interactions are screened out by use of the estimated importance scores. 
Then, we approximate the black-box model by the functional ANOVA model only with selected interactions.
\paragraph{Step 1-1: Screening input features }
Let $\mathcal{V}=\{j: \hat{I}^{(0)}(j)>\tau_0\}$ for a pre-specified positive real number $\tau_0,$ where $\hat{I}^{(0)}(j)$ is the estimate of $\hat{I}^{(0)}(j)$.
The set $\mathcal{V}$ consists of selected input features.
\paragraph{Step 1-2: Screening interactions}
The algorithm is similar to the Apriori algorithm (\cite{srikant1996fast}).
To screen higher order interactions, we sequentially delete unnecessary interactions as follows.
For given $\bold{j}\subset \mathcal{V},$ let $\mathcal{A}(\bold{j})=\{\bold{l}\subset \mathcal{V}: |\bold{l}|=|\bold{j}|-1, \bold{l}\subset \bold{j}\},$ which we call the ancestor set of $\bold{j}$ because any $\bold{l}$ in $\mathcal{A}(\bold{j})$ can be obtained by deleting one entry in $\bold{j}$.
Let $\mathcal{R}$ be the set of candidate interactions.
\begin{itemize}
\item Choose the maximum order $K$ of interactions and let $\mathcal{R}=\{\bold{j}\subset \mathcal{V}: |\bold{j}|=K\}$.
\item For the second and higher interactions, let $\mathcal{C}_1=\{j\in \mathcal{V}: \hat{I}(\{j\}) > \gamma_1\}$ for a pre-specified small positive constant $\gamma_1,$ and we delete all interactions $\bold{j}$ with $|\bold{j}|\ge 2$ and $\bold{j}\cap \mathcal{C}_1^c \ne \emptyset$ from $\mathcal{R}$.
\item For $k\ge 2,$ suppose that $\mathcal{C}_{k-1}$ is given.
To construct $\mathcal{C}_{k},$ we only consider $\bold{j}$ with $|\bold{j}|=k$ such that
all of $\bold{l}\in \mathcal{A}(\bold{j})$ belong to $\mathcal{C}_{k-1}$.
Let $\mathcal{S}_{k}$ be the set of such indices. 
Then, we let $\mathcal{C}_{k}=\{\bold{j}\in \mathcal{S}_{k}: \hat{I}(\bold{j}) > \gamma_k\}$
for a pre-specified positive constant $\gamma_k,$ and delete all interactions $\bold{j}$
with $|\bold{j}|\ge {k+1}$ and $\bold{j}\cap \mathcal{C}_k^c \ne \emptyset$ from $\mathcal{R}$.
\end{itemize}
The interaction screening algorithm is summarized in Algorithm \ref{alg: step1}.
Note that the proposed screening algorithm is exactly the same as the Apriori algorithm
when $\gamma_k$s are all equal and we treat $\bold{j}$ as the item set and $\hat{I}(\bold{j})$ as the support of the item set $\bold{j}$.
As the Apriori algorithm does, the size of $S_k$ decreases fast since it only includes interactions all of whose ancestors have large importance scores.

\begin{algorithm}[h]
   \caption{Interaction screening algorithm}
   \label{alg: step1}   
\begin{algorithmic}
   \STATE {\bfseries Input:} $K$ : the maximum order for interactions , $\gamma_k, k=0,\ldots,K-1$ : thresholds
   \STATE Let $\mathcal{V}=\{j: \hat{I}^{(0)}(j)>\tau_0\}.$
   \STATE Initialize $k=1$
   \STATE Initialize $\mathcal{S}_{1}=\mathcal{V}$ and $\mathcal{R}=\{\bold{j}\subset \mathcal{V}: |\bold{j}|=K\}$
   \WHILE{$k \leq K$}
   \STATE $\mathcal{C}_{k} = \{ \bold{j} \in \mathcal{S}_{k} : \hat{I}(\bold{j}) > \gamma_k \}$
   \STATE Delete all $\bold{j}'$ with $|\bold{j}'|> k$ and $\bold{j}' \cap \mathcal{C}_k^c \ne \emptyset$
   from $\mathcal{R}.$
   \STATE $\mathcal{S}_{k+1} =\{\bold{j} \subset \mathcal{V}: |\bold{j}|=k+1, \mathcal{A}(\bold{j}) \subset \mathcal{C}_k\}.$  
   
  \STATE $k \xleftarrow{} k+1$
   \ENDWHILE
\end{algorithmic}
\end{algorithm}

\paragraph{Step 2: Learning the function ANOVA model only with selected interactions}

After obtaining $\mathcal{R}$ in Step 1, we consider the following partial functional ANOVA model:
\begin{align}
\begin{split}
\displaystyle 
f_{\mathcal{R}}(\bold{x})=\beta_{0} + &\sum_{k=1}^{K} \sum_{\bold{j} \in \mathcal{R}_k} 
f_{\bold{j}}(\bold{x}_{\bold{j}}),
\end{split}
\end{align}
where $\mathcal{R}_k=\mathcal{R}\cap \{\bold{j}\subset [p]: |\bold{j}|=k\}$.
We estimate $f_{\mathcal{R}}$ by minimizing $\sum_{i=1}^n (f(\bold{x}^{i})-f_{\mathcal{R}}(\bold{x}^{i}))^2$.

Possible algorithms for estimating $f_{\mathcal{R}}$ would be smoothing spline (\cite{gu2013smoothing}) and NAM (\cite{agarwal2020neural}). 
Smoothing spline is computationally demanding when the size of data is large, while
NAM only includes the main effects.
In this paper, we use a modified version of NAM, so-called Neural Interaction Model (NIM), whose details are given in Section \ref{app: NIM details} of Appendix. 

One may argue that we could use NIM for the original functional ANOVA model without interaction screening. 
This approach, however, is computationally prohibitive since the number of interactions 
becomes too large even when $p$ is mildly large. For example, when $p=50,$ the number of all possible third-order interactions becomes 20,875.
{\it For including higher order interactions into the model, interaction screening is a must!}
See Section \ref{app:compute_complex} of Appendix for numerical results.

\subsection{Remarks about computational complexity}\label{sec:compute_complexity}
Calculating $\widehat{D}_{\bold{j}} f_0(\mathbf{x})$ needs
$2^{|\bold{j}|}$ many computations and thus
computational complexity for calculating $\hat{I}(\bold{j})$
is proportional to $2^{|\bold{j}|} n^2,$ where $n^2$ comes from the computations of 
$\mathbb{E}_{\mathbf{X}_{\bold{j}}'\sim \mathbb{P}_{\bold{j}}}$
and $\operatorname{Var}{\mathbf{X}_{\bold{j}^c}'\sim \mathbb{P}_{\bold{j}^c}}$.
Hence, The total computational complexity of interaction screening is proportional to
$\sum_{k=1}^{K-1} |\mathcal{S}_k| 2^{k} n^{2}.$ 
 Note that $|\mathcal{S}_k|$ is data-dependent and is usually expected to be small for large $k$ because not many higher order interactions are significant. See Section \ref{app:compute_interact_experiment} of Appendix
 for numerical evidences.

\section{Experiments}
\label{sec:experiment}

We conduct experiments to evaluate the performance of Meta-ANOVA, focusing on three aspects: 1) how effectively the proposed interaction screening algorithm can find the true signal interactions, 2) how well the model learned
by Meta-ANOVA approximates the baseline black-box model, and 3) how  useful the approximated model is in view of XAI.
The evaluation is carried out by
analyzing synthetic as well as real datasets. Details of using Meta-ANOVA including the choice of the thresholds $\gamma_k$ and the bandwidths $h_{k,n}$ are presented in Section \ref{app: details} of Appendix.

\subsection{Interaction detection} \label{sec: syn}

We investigate how well Meta-ANOVA selects signal interactions 
by analyzing synthetic data.
For this purpose, we consider 10 synthetic regression models, as shown in Section \ref{app: syn_details} of  Appendix which are used in (\cite{tsang2017detecting, liu2020detecting}).
We generate data from each of the synthetic regression models (without adding noises) where the input features are generated from the uniform distribution. 
We apply the interaction screening algorithm of Meta-ANOVA to the simulated data and assess how well Meta-ANOVA screens signal interactions compared to other competitors. 
We only consider selecting second-order interactions.
See Section \ref{app: score interaction} of Appendix for measuring the importance of each interaction screened by Meta-ANOVA.


\begin{table}[H]
\centering
\caption{Comparison of the AUROCs for selecting second-order interactions using Meta-ANOVA and other competitors on synthetic regression models.}
\label{table:pair_auc}
\begin{tabular}{@{}lccccc@{}}
\toprule
        & RuleFit & AG    & NID   & PID   & Meta-ANOVA \\\midrule
        
$F_1$    & 0.754   & 1.000 & 0.985 & 0.986 & 1.000 \\
$F_2$    & 0.698   & 0.880 & 0.776 & 0.804 & 0.866 \\
$F_3$    & 0.815   & 1.000 & 1.000 & 1.000 & 1.000 \\
$F_4$    & 0.689   & 0.999 & 0.916 & 0.935 & 1.000 \\
$F_5$    & 0.797   & 0.670 & 0.997 & 1.000 & 0.894 \\
$F_6$    & 0.811   & 0.640 & 0.999 & 1.000 & 1.000 \\
$F_7$    & 0.666   & 0.810 & 0.880 & 0.888 & 0.759 \\
$F_8$    & 0.946   & 0.937 & 1.000 & 1.000 & 0.947 \\
$F_9$    & 0.584   & 0.808 & 0.968 & 0.972 & 0.752 \\
$F_{10}$ & 0.876   & 1.000 & 0.989 & 0.987 & 1.000 \\\midrule
Average  & 0.764   & 0.870 & 0.951 & 0.957 & 0.922 \\\bottomrule
\end{tabular}
\end{table}

As baselines for interaction selection, we consider four methods
RuleFit (\cite{friedman2008predictive}), Additive Groves (AG, \cite{sorokina2008detecting}), NID (\cite{tsang2017detecting}), and PID (\cite{liu2020detecting}). 
Note that RuleFit and AG select interactions from a tree based model, and NID and PID do from a deep neural network, while Meta-ANOVA is model-agnostic. 

Table \ref{table:pair_auc} summarizes the Area Under ROC curve (AUROC) values 
based on the ranks of the importance of second-order interactions obtained by each selection method.
The results except those of Meta-ANOVA are copied from \cite{liu2020detecting}.
For Meta-ANOVA, we use a DNN for the baseline black-box model. 
Details of the experiment including the 10 synthetic regression models,
are given  in Section \ref{app: syn_details} of Appendix.

Meta-ANOVA is superior to RuleFit and AG, but is slightly inferior to NID and PID.
The inferior performance of Meta-ANOVA compared to NID and PID is, however, not surprising
since NID and PID are DNN specific methods while Meta-ANOVA is model-agnostic.
We think that these results amply support that Meta-ANOVA is an useful model-agnostic interaction screening algorithm.

\subsection{Prediction performance}

We analyze 5 benchmark real datasets: \texttt{Calhousing} (\cite{scikit-learn}), \texttt{Letter} (\cite{frey1991letter}), \texttt{German credit} (\cite{gromping2019south}), \texttt{Online news} (\cite{misc_online_news_popularity_332}), and \texttt{Abalone} (\cite{misc_abalone_1}). 
Details of each dataset are described in Section \ref{app: NIM detalis for real} of Appendix. 
For all datasets, we split data into train/validation/test with the ratio of 70/10/20. 
Also, continuous input features are normalized using the minmax scaler and
categorical input features are preprocessed using the one-hot encoding.
The target variable for regression problem is normalized using the standard scaler. 
All reported results are the averages (and standard errors) of results obtained from 10 random splits of data.

To measure prediction performance,
Mean Squared Error (MSE) is used for regression and AUROC is used for classification.
For the baseline black-box model, we consider the following three algorithms:
1) Deep Neural Network (DNN) with four hidden layers (140-100-60-20), 2) Extreme Gradient Boosting (XGB, \cite{Chen:2016:XST:2939672.2939785}),
and 3) Random Forest (RF, \cite{breiman2001random}). We choose the most accurate model for each dataset as the baseline black-box model.
See Section \ref{app: NIM detalis for real} of Appendix for the comparisons of the three black-box models.
In addition, we choose $K$ among 2,3, and 4, which yields the best result.
The results for other $K$s are presented in Section \ref{app: order NIM} of Appendix.
\vskip -0.1in
\begin{table}[H]
    \centering
    \caption{Prediction performances of the baseline black-box models and corresponding Meta-ANOVA models: The models in the parenthesis are the selected baseline black-box models.}
    \label{Table : perfor-NIM}
    \begin{tabular}{cccccr}
    \toprule
    Dataset & Measure & Baseline &Meta-ANOVA & Max order  \\
    \midrule
    \midrule
    \texttt{Calhousing} & MSE $\downarrow$ &0.164 (XGB) & 0.165 & 4\\
    \texttt{Abalone}    &MSE $\downarrow$&0.432 (DNN) & 0.427 & 4 \\ 
    \midrule
    \texttt{German credit} & AUROC $\uparrow$&0.787 (DNN) & 0.778 &  2\\
    \texttt{Online} & AUROC $\uparrow$& 0.723 (RF) & 0.720 & 4\\
    \texttt{Letter} & AUROC $\uparrow$& 0.996 (RF) & 0.994 & 4\\
    \bottomrule
    \end{tabular}
\end{table}

The results are presented in Table \ref{Table : perfor-NIM}.
It is obvious that the losses of performance due to the approximation
by Meta-ANOVA are minimal. In particular, for the \texttt{Abalone} dataset, even the approximated model performs better.
These results indicate that Meta-ANOVA approximates given black-box models closely.

\subsection{Interpretability}

SHAP (\cite{lundberg2017unified}) is one of the most popularly used methods for XAI, which gives the importance measures of each input feature vector (local SHAP) and each input feature (global SHAP).
In this section, we compare the global interpretations of Meta-ANOVA and global SHAP (\cite{molnar2020interpretable}).
The results for local interpretation are presented in Appendix \ref{app: shap}.

Table \ref{table: shap,meta} compares the feature importance measures of the 5 most important features in \texttt{Calhousing} dataset selected by Global SHAP and Meta-ANOVA.
For global SHAP, we use the ``\texttt{shap}'' python package.
For the global importance measure of each input feature for Meta-ANOVA, see
Section \ref{app: shap} of Appendix.
We observe that the two results are similar. In particular, the three most important input features are the same.
\begin{table}[H]
\small
\centering
\caption{Feature importance measures of the 5 most important features 
of global SHAP and Meta-ANOVA}
\label{table: shap,meta}
\begin{tabular}{@{}clccccc@{}}
\toprule
\multirow{2}{*}{\textbf{Global SHAP}} & \textbf{Selected features} & Latitude  & Longitude & MedInc & AveOcc  & AveRoom  \\ \cmidrule(l){2-7} 
                             & \textbf{Feature importances}            & 1.000     & 0.861     & 0.759  & 0.379   &  0.211  \\ \midrule\midrule
\multirow{2}{*}{\textbf{Meta-ANOVA}}  & \textbf{Selected features} & Longitude & Latitude  & MedInc & AveRoom & HouseAge  \\ \cmidrule(l){2-7} 
                             & \textbf{Feature importances}            & 1.000     &  0.962    & 0.383  & 0.009   &  0.004  \\ \bottomrule
\end{tabular}
\end{table}

\subsection{Ablation studies and Application to large-sized models}
The results of various ablation studies for the large complex models (i.e., TabTransformer (\cite{huang2020tabtransformer})) are presented in Section \ref{app: ablation} of Appendix due to the page limitation.

Furthermore, we additionally demonstrate that our Meta-ANOVA is applicable to interpret large-sized models for image and text data. We considered ResNet-18 (\cite{he2016deep}) and SST2-DistilBERT (\cite{sanh2019distilbert}) trained on \texttt{CelebA} (\cite{celeba}) and \texttt{GLUE-SST2} (\cite{wang2018glue}) as baseline black-box models, respectively. Detailed results of experiments are presented in Section \ref{app:image_data} and \ref{app:text_data} of Appendix.

\section{Conclusion}
\label{sec:conclusion}

In this paper, we have proposed a new post-processing interpretation method called Meta-ANOVA.
The novel contribution of Meta-ANOVA is the interaction screening algorithm, which is computationally
efficient and theoretically sound. In addition, Meta-ANOVA transfers the baseline black-box model
to a white-box functional ANOVA model, making its interpretation consistent and transparent.
Thus, Meta-ANOVA can be used as an auxiliary tool to reconfirm
the validity of results of existing XAI methods.

Meta-ANOVA could be used as an alternative algorithm to fit the functional ANOVA model from data.
A standard approach is to estimate the functional ANOVA model by minimizing the loss function.
Instead, we could deliberately fit a  baseline black-box such as ensembles or DNNs,
ignoring the structure of functional ANOVA from data, and transfer it to the functional ANOVA model. 
At this point, it is not obvious what the advantages of this two stage process are, but
we will pursue it in the near future.

We use NIM, which is a straightforward extension of NAM (\cite{agarwal2020neural}).
Recently, various neural networks for the functional ANOVA model, including
NBM (Neural Basis Model, \cite{radenovic2022neural}) and NODE-GAM (\cite{chang2021node}), have been developed. These algorithms can be used for Meta-ANOVA without any modification.

\vspace{-0.1in}
\paragraph{Broader Impacts:}This paper presents work whose goal is to advance the field of interpreting the machine learning models. Whilen there are many potential societal consequences of our work, we do not believe any are tied to specific broader impacts or risks.



\bibliographystyle{unsrt}  
\bibliography{references}  

\begin{thebibliography}{10}

\bibitem{he2016deep}
Kaiming He, Xiangyu Zhang, Shaoqing Ren, and Jian Sun.
\newblock Deep residual learning for image recognition.
\newblock In {\em Proceedings of the IEEE conference on computer vision and pattern recognition}, pages 770--778, 2016.

\bibitem{radford2019language}
Alec Radford, Jeffrey Wu, Rewon Child, David Luan, Dario Amodei, Ilya Sutskever, et~al.
\newblock Language models are unsupervised multitask learners.
\newblock {\em OpenAI blog}, 1(8):9, 2019.

\bibitem{shen2017deep}
Dinggang Shen, Guorong Wu, and Heung-Il Suk.
\newblock Deep learning in medical image analysis.
\newblock {\em Annual review of biomedical engineering}, 19:221--248, 2017.

\bibitem{devlin2018bert}
Jacob Devlin, Ming-Wei Chang, Kenton Lee, and Kristina Toutanova.
\newblock Bert: Pre-training of deep bidirectional transformers for language understanding.
\newblock {\em arXiv preprint arXiv:1810.04805}, 2018.

\bibitem{chouiekh2018convnets}
Alae Chouiekh and EL~Hassane Ibn~EL Haj.
\newblock Convnets for fraud detection analysis.
\newblock {\em Procedia Computer Science}, 127:133--138, 2018.

\bibitem{sharif2014cnn}
Ali Sharif~Razavian, Hossein Azizpour, Josephine Sullivan, and Stefan Carlsson.
\newblock Cnn features off-the-shelf: an astounding baseline for recognition.
\newblock In {\em Proceedings of the IEEE conference on computer vision and pattern recognition workshops}, pages 806--813, 2014.

\bibitem{webb2018deep}
Sarah Webb et~al.
\newblock Deep learning for biology.
\newblock {\em Nature}, 554(7693):555--557, 2018.

\bibitem{beck2020overview}
Christian Beck, Martin Hutzenthaler, Arnulf Jentzen, and Benno Kuckuck.
\newblock An overview on deep learning-based approximation methods for partial differential equations.
\newblock {\em arXiv preprint arXiv:2012.12348}, 2020.

\bibitem{gu2013smoothing}
Chong Gu.
\newblock {\em Smoothing spline ANOVA models}, volume 297.
\newblock Springer, 2013.

\bibitem{caru2019purifying}
Benjamin Lengerich, Sarah Tan, Chun-Hao Chang, Giels Hooker, and Rich Caruana.
\newblock Purifying interaction effects with the functional anova: An efficient algorithm for recovering identifiable additive models.
\newblock {\em International Conference on Artificial Intelligence and Statistics}, pages 2402--2412, 2020.

\bibitem{chong1993ssanova}
Chong Gu and Grace Wahba.
\newblock Smoothing spline anova with component-wise bayesian confidence intervals.
\newblock {\em Journal of Computational and Graphical Statistics}, 2(1):97--117, 1993.

\bibitem{kim2009boosting}
Jinseog Kim, Yongdai Kim, Yuwon Kim, Sunghoon Kwon, and Sangin Lee.
\newblock Boosting on the functional anova decomposition.
\newblock {\em Statistics and Its Interface}, 2(3):361--368, 2009.

\bibitem{lin2006component}
Yi~Lin and Hao~Helen Zhang.
\newblock {Component selection and smoothing in multivariate nonparametric regression}.
\newblock {\em The Annals of Statistics}, 34(5):2272 -- 2297, 2006.

\bibitem{fan2008sure}
Jianqing Fan and Jinchi Lv.
\newblock Sure independence screening for ultrahigh dimensional feature space.
\newblock {\em Journal of the Royal Statistical Society Series B: Statistical Methodology}, 70(5):849--911, 2008.

\bibitem{tibshirani1996regression}
Robert Tibshirani.
\newblock Regression shrinkage and selection via the lasso.
\newblock {\em Journal of the Royal Statistical Society Series B: Statistical Methodology}, 58(1):267--288, 1996.

\bibitem{wang2020simple}
Gao Wang, Abhishek Sarkar, Peter Carbonetto, and Matthew Stephens.
\newblock A simple new approach to variable selection in regression, with application to genetic fine mapping.
\newblock {\em Journal of the Royal Statistical Society Series B: Statistical Methodology}, 82(5):1273--1300, 2020.

\bibitem{linardatos2020explainable}
Pantelis Linardatos, Vasilis Papastefanopoulos, and Sotiris Kotsiantis.
\newblock Explainable ai: A review of machine learning interpretability methods.
\newblock {\em Entropy}, 23(1):18, 2020.

\bibitem{lipton2018mythos}
Zachary~C Lipton.
\newblock The mythos of model interpretability: In machine learning, the concept of interpretability is both important and slippery.
\newblock {\em Queue}, 16(3):31--57, 2018.

\bibitem{doshi2017towards}
Finale Doshi-Velez and Been Kim.
\newblock Towards a rigorous science of interpretable machine learning.
\newblock {\em arXiv preprint arXiv:1702.08608}, 2017.

\bibitem{gilpin2018explaining}
Leilani~H Gilpin, David Bau, Ben~Z Yuan, Ayesha Bajwa, Michael Specter, and Lalana Kagal.
\newblock Explaining explanations: An overview of interpretability of machine learning.
\newblock In {\em 2018 IEEE 5th International Conference on data science and advanced analytics (DSAA)}, pages 80--89. IEEE, 2018.

\bibitem{montavon2018methods}
Gr{\'e}goire Montavon, Wojciech Samek, and Klaus-Robert M{\"u}ller.
\newblock Methods for interpreting and understanding deep neural networks.
\newblock {\em Digital Signal Processing}, 73:1--15, 2018.

\bibitem{linardatos2021explainable}
Pantelis Linardatos, Vasilis Papastefanopoulos, and Sotiris Kotsiantis.
\newblock Explainable ai: A review of machine learning interpretability methods.
\newblock {\em Entropy}, 23(1):18, 2021.

\bibitem{melis2018towards}
David~Alvarez Melis and Tommi Jaakkola.
\newblock Towards robust interpretability with self-explaining neural networks.
\newblock In {\em Advances in Neural Information Processing Systems}, pages 7775--7784, 2018.

\bibitem{zhang2018interpretable}
Quanshi Zhang, Ying Nian~Wu, and Song-Chun Zhu.
\newblock Interpretable convolutional neural networks.
\newblock In {\em Proceedings of the IEEE Conference on Computer Vision and Pattern Recognition}, pages 8827--8836, 2018.

\bibitem{li2018deep}
Oscar Li, Hao Liu, Chaofan Chen, and Cynthia Rudin.
\newblock Deep learning for case-based reasoning through prototypes: A neural network that explains its predictions.
\newblock In {\em Proceedings of the AAAI Conference on Artificial Intelligence}, volume~32, Apr. 2018.

\bibitem{fukui2019attention}
Hiroshi Fukui, Tsubasa Hirakawa, Takayoshi Yamashita, and Hironobu Fujiyoshi.
\newblock Attention branch network: Learning of attention mechanism for visual explanation.
\newblock In {\em Proceedings of the IEEE/CVF Conference on Computer Vision and Pattern Recognition}, pages 10705--10714, 2019.

\bibitem{wang2021self}
Yipei Wang and Xiaoqian Wang.
\newblock Self-interpretable model with transformation equivariant interpretation.
\newblock {\em Advances in Neural Information Processing Systems}, 34, 2021.

\bibitem{agarwal2020neural}
Rishabh Agarwal, Nicholas Frosst, Xuezhou Zhang, Rich Caruana, and Geoffrey~E Hinton.
\newblock Neural additive models: Interpretable machine learning with neural nets.
\newblock {\em Advances in Neural Information Processing Systems}, 34, 2021.

\bibitem{hastie2017generalized}
Trevor~J Hastie.
\newblock Generalized additive models.
\newblock In {\em Statistical models in S}, pages 249--307. Routledge, 2017.

\bibitem{radenovic2022neural}
Filip Radenovic, Abhimanyu Dubey, and Dhruv Mahajan.
\newblock Neural basis models for interpretability.
\newblock {\em Advances in Neural Information Processing Systems}, 35:8414--8426, 2022.

\bibitem{chang2021node}
Chun-Hao Chang, Rich Caruana, and Anna Goldenberg.
\newblock Node-gam: Neural generalized additive model for interpretable deep learning.
\newblock {\em arXiv preprint arXiv:2106.01613}, 2021.

\bibitem{zeiler2014visualizing}
Matthew~D Zeiler and Rob Fergus.
\newblock Visualizing and understanding convolutional networks.
\newblock In {\em European conference on computer vision}, pages 818--833. Springer, 2014.

\bibitem{simonyan2013deep}
Karen Simonyan, Andrea Vedaldi, and Andrew Zisserman.
\newblock Deep inside convolutional networks: Visualising image classification models and saliency maps.
\newblock {\em arXiv preprint arXiv:1312.6034}, 2013.

\bibitem{sundararajan2017axiomatic}
Mukund Sundararajan, Ankur Taly, and Qiqi Yan.
\newblock Axiomatic attribution for deep networks.
\newblock In {\em International Conference on Machine Learning}, pages 3319--3328. PMLR, 2017.

\bibitem{shrikumar2017learning}
Avanti Shrikumar, Peyton Greenside, and Anshul Kundaje.
\newblock Learning important features through propagating activation differences.
\newblock In {\em International Conference on Machine Learning}, pages 3145--3153. PMLR, 2017.

\bibitem{zhou2016learning}
Bolei Zhou, Aditya Khosla, Agata Lapedriza, Aude Oliva, and Antonio Torralba.
\newblock Learning deep features for discriminative localization.
\newblock In {\em Proceedings of the IEEE conference on computer vision and pattern recognition}, pages 2921--2929, 2016.

\bibitem{selvaraju2017grad}
Ramprasaath~R Selvaraju, Michael Cogswell, Abhishek Das, Ramakrishna Vedantam, Devi Parikh, and Dhruv Batra.
\newblock Grad-cam: Visual explanations from deep networks via gradient-based localization.
\newblock In {\em Proceedings of the IEEE international conference on computer vision}, pages 618--626, 2017.

\bibitem{chattopadhay2018grad}
Aditya Chattopadhay, Anirban Sarkar, Prantik Howlader, and Vineeth~N Balasubramanian.
\newblock Grad-cam++: Generalized gradient-based visual explanations for deep convolutional networks.
\newblock In {\em 2018 IEEE Winter Conference on Applications of Computer Vision (WACV)}, pages 839--847. IEEE, 2018.

\bibitem{wang2020score}
Haofan Wang, Zifan Wang, Mengnan Du, Fan Yang, Zijian Zhang, Sirui Ding, Piotr Mardziel, and Xia Hu.
\newblock Score-cam: Score-weighted visual explanations for convolutional neural networks.
\newblock In {\em Proceedings of the IEEE/CVF conference on computer vision and pattern recognition workshops}, pages 24--25, 2020.

\bibitem{tsang2017detecting}
Michael Tsang, Dehua Cheng, and Yan Liu.
\newblock Detecting statistical interactions from neural network weights.
\newblock {\em arXiv preprint arXiv:1705.04977}, 2017.

\bibitem{liu2020detecting}
Zirui Liu, Qingquan Song, Kaixiong Zhou, Ting-Hsiang Wang, Ying Shan, and Xia Hu.
\newblock Detecting interactions from neural networks via topological analysis.
\newblock {\em Advances in Neural Information Processing Systems}, 33, 2020.

\bibitem{ribeiro2016should}
Marco~Tulio Ribeiro, Sameer Singh, and Carlos Guestrin.
\newblock ``why should i trust you?" explaining the predictions of any classifier.
\newblock In {\em Proceedings of the 22nd ACM SIGKDD international conference on knowledge discovery and data mining}, pages 1135--1144, 2016.

\bibitem{zafar2019dlime}
Muhammad~Rehman Zafar and Naimul~Mefraz Khan.
\newblock Dlime: A deterministic local interpretable model-agnostic explanations approach for computer-aided diagnosis systems.
\newblock {\em arXiv preprint arXiv:1906.10263}, 2019.

\bibitem{shapley1953value}
Lloyd~S Shapley et~al.
\newblock A value for n-person games.
\newblock 1953.

\bibitem{lundberg2017unified}
Scott Lundberg and Su-In Lee.
\newblock A unified approach to interpreting model predictions.
\newblock {\em arXiv preprint arXiv:1705.07874}, 2017.

\bibitem{masoomi2021explanations}
Aria Masoomi, Davin Hill, Zhonghui Xu, Craig~P Hersh, Edwin~K Silverman, Peter~J Castaldi, Stratis Ioannidis, and Jennifer Dy.
\newblock Explanations of black-box models based on directional feature interactions.
\newblock In {\em International Conference on Learning Representations}, 2021.

\bibitem{tsai2023faith}
Che-Ping Tsai, Chih-Kuan Yeh, and Pradeep Ravikumar.
\newblock Faith-shap: The faithful shapley interaction index.
\newblock {\em Journal of Machine Learning Research}, 24(94):1--42, 2023.

\bibitem{lundstrom2023unifying}
Daniel Lundstrom and Meisam Razaviyayn.
\newblock A unifying framework to the analysis of interaction methods using synergy functions.
\newblock In {\em International Conference on Machine Learning}, pages 23005--23032. PMLR, 2023.

\bibitem{srikant1996fast}
Rakesh Agrawal and Ramakrishnan Srikant.
\newblock {\em Fast algorithms for mining association rules and sequential patterns}.
\newblock The University of Wisconsin-Madison, 1996.

\bibitem{stone1977consistent}
Charles~J Stone.
\newblock Consistent nonparametric regression.
\newblock {\em The annals of statistics}, pages 595--620, 1977.

\bibitem{tsybakov1986robust}
Aleksandr~Borisovich Tsybakov.
\newblock Robust reconstruction of functions by the local-approximation method.
\newblock {\em Problemy Peredachi Informatsii}, 22(2):69--84, 1986.

\bibitem{fan1997local}
Jianqing Fan, Theo Gasser, Ir{\`e}ne Gijbels, Michael Brockmann, and Joachim Engel.
\newblock Local polynomial regression: Optimal kernels and asymptotic minimax efficiency.
\newblock {\em Annals of the Institute of Statistical Mathematics}, 49:79--99, 1997.

\bibitem{imaizumi2023sup}
Masaaki Imaizumi.
\newblock Sup-norm convergence of deep neural network estimator for nonparametric regression by adversarial training.
\newblock {\em arXiv preprint arXiv:2307.04042}, 2023.

\bibitem{saltelli2010variance}
Andrea Saltelli, Paola Annoni, Ivano Azzini, Francesca Campolongo, Marco Ratto, and Stefano Tarantola.
\newblock Variance based sensitivity analysis of model output. design and estimator for the total sensitivity index.
\newblock {\em Computer physics communications}, 181(2):259--270, 2010.

\bibitem{friedman2008predictive}
Jerome~H Friedman and Bogdan~E Popescu.
\newblock Predictive learning via rule ensembles.
\newblock {\em The Annals of Applied Statistics}, 2(3):916--954, 2008.

\bibitem{sorokina2008detecting}
Daria Sorokina, Rich Caruana, Mirek Riedewald, and Daniel Fink.
\newblock Detecting statistical interactions with additive groves of trees.
\newblock In {\em Proceedings of the 25th international conference on Machine learning}, pages 1000--1007, 2008.

\bibitem{scikit-learn}
F.~Pedregosa, G.~Varoquaux, A.~Gramfort, V.~Michel, B.~Thirion, O.~Grisel, M.~Blondel, P.~Prettenhofer, R.~Weiss, V.~Dubourg, J.~Vanderplas, A.~Passos, D.~Cournapeau, M.~Brucher, M.~Perrot, and E.~Duchesnay.
\newblock Scikit-learn: Machine learning in {P}ython.
\newblock {\em Journal of Machine Learning Research}, 12:2825--2830, 2011.

\bibitem{frey1991letter}
Peter~W Frey and David~J Slate.
\newblock Letter recognition using holland-style adaptive classifiers.
\newblock {\em Machine learning}, 6(2):161--182, 1991.

\bibitem{gromping2019south}
Ulrike Gr{\"o}mping.
\newblock South german credit data: Correcting a widely used data set.
\newblock {\em Rep. Math., Phys. Chem., Berlin, Germany, Tech. Rep}, 4:2019, 2019.

\bibitem{misc_online_news_popularity_332}
Kelwin Fernandes, Pedro Vinagre, Paulo Cortez, and Pedro Sernadela.
\newblock {Online News Popularity}.
\newblock UCI Machine Learning Repository, 2015.
\newblock {DOI}: https://doi.org/10.24432/C5NS3V.

\bibitem{misc_abalone_1}
Warwick Nash, Tracy Sellers, Simon Talbot, Andrew Cawthorn, and Wes Ford.
\newblock {Abalone}.
\newblock UCI Machine Learning Repository, 1995.
\newblock {DOI}: https://doi.org/10.24432/C55C7W.

\bibitem{Chen:2016:XST:2939672.2939785}
Tianqi Chen and Carlos Guestrin.
\newblock {XGBoost}: A scalable tree boosting system.
\newblock In {\em Proceedings of the 22nd ACM SIGKDD International Conference on Knowledge Discovery and Data Mining}, KDD '16, pages 785--794, New York, NY, USA, 2016. ACM.

\bibitem{breiman2001random}
Leo Breiman.
\newblock Random forests.
\newblock {\em Machine learning}, 45:5--32, 2001.

\bibitem{molnar2020interpretable}
Christoph Molnar.
\newblock {\em Interpretable machine learning}.
\newblock Lulu. com, 2020.

\bibitem{huang2020tabtransformer}
Xin Huang, Ashish Khetan, Milan Cvitkovic, and Zohar Karnin.
\newblock Tabtransformer: Tabular data modeling using contextual embeddings.
\newblock {\em arXiv preprint arXiv:2012.06678}, 2020.

\bibitem{sanh2019distilbert}
Victor Sanh, L~Debut, J~Chaumond, and T~Wolf.
\newblock Distilbert, a distilled version of bert: Smaller, faster, cheaper and lighter. arxiv 2019.
\newblock {\em arXiv preprint arXiv:1910.01108}, 2019.

\bibitem{celeba}
Ziwei Liu, Ping Luo, Xiaogang Wang, and Xiaoou Tang.
\newblock Deep learning face attributes in the wild.
\newblock In {\em Proceedings of the IEEE international conference on computer vision}, pages 3730--3738, 2015.

\bibitem{wang2018glue}
Alex Wang, Amanpreet Singh, Julian Michael, Felix Hill, Omer Levy, and Samuel~R Bowman.
\newblock Glue: A multi-task benchmark and analysis platform for natural language understanding.
\newblock {\em arXiv preprint arXiv:1804.07461}, 2018.

\bibitem{hooker2007ganova}
Giles Hooker.
\newblock Generalized functional anova diagnostics for high-dimensional functions of dependent variable.
\newblock {\em Journal of Computational and Graphical Statistics}, 16(3):709--732, 2007.

\bibitem{molnar2022interpretable2nd}
Christoph Molnar.
\newblock {\em Interpretable machine learning}.
\newblock Lulu. com, second edition, 2022.

\bibitem{koh2020concept}
Pang~Wei Koh, Thao Nguyen, Yew~Siang Tang, Stephen Mussmann, Emma Pierson, Been Kim, and Percy Liang.
\newblock Concept bottleneck models.
\newblock In {\em International conference on machine learning}, pages 5338--5348. PMLR, 2020.

\end{thebibliography}

\newpage
\appendix
\onecolumn
\section{Proofs}\label{app:proofs}
\subsection{Proof of Theorem \ref{thm:conti}} \label{proof:conti}

Before going further, we introduce the definition of statistical interaction from \cite{sorokina2008detecting}.
{\definition{Statistical Interaction:}\label{stat_int}
A function $f(\bold{x})$ does not possesses interaction $\bold{j}=(j_1,\cdots,j_k)$ if it can be expressed as the sum of $k$ (or fewer) functions, $f_{\text{\textbackslash} j_1},\cdots,f_{\text{\textbackslash}j_k}$ as follows:
$$
f(\bold{x}) = \sum_{\ell=1}^k f_{\text{\textbackslash}j_\ell}(x_1,\cdots,x_{j_{\ell}-1},x_{j_{\ell}+1},\cdots, x_p)
$$
}

\textit{(Proof of Theorem \ref{thm:conti})}

Note that for $\bold{j}\subset [p]$, we rewrite $f_0$ as:
\begin{align*}
f_0(\bold{x})=\beta_{0}+\sum_{\bold{j}' <\bold{j}}G_{\bold{j}'\bold{j}}(\bold{x})+G_{\bold{j}\bold{j}}(\bold{x})+\sum_{\bold{j}'' < \bold{j}^{c}}f_{0,\bold{j}''}(\bold{x}_{\bold{j}''}),
\end{align*}
where $G_{\bold{j}'\bold{j}}(\bold{x})=f_{0,\bold{j}'}(\bold{x}_{\bold{j}'})+\sum_{\bold{j}_1\subset\bold{j}^{c}}f_{0,\bold{j}'\cup\bold{j}_1}(\bold{x}_{\bold{j}'\cup\bold{j}_1}).$

`Only if' part : \\
$\newline$
For a given $\bold{j} \subset [p]$, suppose that $f_{0,\bold{j}'} =  0$ for all $\bold{j}'>\bold{j}.$ Then, we can write 
\begin{align*}
f_0(\bold{x})=\beta_0 + \sum_{\bold{j}'<\bold{j}}G_{\bold{j}'\bold{j}}(\bold{x}) + f_{0,\bold{j}}(\bold{x}_{\bold{j}})+ \sum_{\bold{j}'' < \bold{j}^{c}}f_{0,\bold{j}''}(\bold{x}_{\bold{j}''}),
\end{align*}
and $D_{\bold{j}}f_0(\bold{x})=D_{\bold{j}}f_{0,\bold{j}}(\bold{x}_{\bold{j}})$ which does not depend on $\bold{x}_{\bold{j}^c}$. 
Therefore $\operatorname{Var}_{\bold{X}_{\bold{j}^c}' \sim \mathbb{P}_{\bold{j}^c}}\left\{ D_{\bold{j}}f_0(\bold{x}_{\bold{j}}',\bold{X}_{\bold{j}^c}')\right\}=0.$ \\
$\newline$
`If' part : \\
$\newline$
For a given $\bold{j} = \{j_1,\cdots,j_k\} \subset [p]$, $\operatorname{Var}_{\bold{X}_{\bold{j}^c}' \sim \mathbb{P}_{\bold{j}^c}}\left\{ D_{\bold{j}}f_0(\bold{x}_{\bold{j}}',\bold{X}_{\bold{j}^c}')\right\}=0$ means that 
$ D_{\bold{j}}f_0(\bold{x})$ does not depend on $\bold{x}_{\bold{j}^c}$. We denote  $\phi(\bold{x}_{\bold{j}}):=D_{\bold{j}} f_0(\bold{x})$. Using the fundamental theorem of calculus, $f$ can be decomposed as follows:
\begin{align}
f_0(\bold{x}_{\bold{j}},\bold{x}_{\bold{j}^c}) & = \int_{\prod_{j\in\bold{j}}[0,x_{j}]}D_{\bold{j}} f_0(\bold{x}_{\bold{j}},\bold{x}_{\bold{j}^c}) d\bold{x}_{\bold{j}} + f_0(\bold{0}_{\bold{j}},\bold{x}_{\bold{j}^c})\nonumber\\
& = \int_{\prod_{j\in\bold{j}}[0,x_j]} \phi(\bold{x}_{\bold{j}}) d\bold{x}_{\bold{j}} + f_0(\bold{0}_{\bold{j}},\bold{x}_{\bold{j}^c})\nonumber\\
&:= h_{0,\bold{j}}(\bold{x}_{\bold{j}}) + h_{0,\bold{j}^c}(\bold{x}_{\bold{j}^c})\quad\quad\quad\quad\quad\quad\quad\quad \forall (\bold{x}_{\bold{j}}, \bold{x}_{\bold{j}^c}) \in \mathcal{X}, \label{decompose}
\end{align}
where the last equation (\ref{decompose}) indicates that $f_0$ can be decomposed as the sum of two functions, $h_{0,\bold{j}}$ and $h_{0,\bold{j}^c}$.
Therefore, by Definition \ref{stat_int}, $f_0$ possesses no interaction $\bold{j}\cup\{j'\}$ for any $j'\in\bold{j}^c$, leading to conclusion that $f_{0,\bold{j}'} = 0$ for all $\bold{j}'>\bold{j}$. $\square$

\subsection{Proof of Theorem \ref{thm:deriv_est}} \label{proof:deriv_est}

Since $D_jf_0$ is the partial derivative with respect to the input feature $x_j$
while the other features are fixed,
we treat $f$ and $f_0$ as univariate functions of $x=x_j$.
For notational simplicity, we drop ${\bold x}_j$ and the subscript $j$ in the notations.
That is, we write simply $D_j f(\bold{x})= Df(x).$

We can bound the $L^{\infty}$-norm of $\widehat{D} f_0-D f_0$ as follows:
\begin{align}
\left\|\widehat{D} f_0(x) - D f_0(x)\right\|_{\infty} & = \left\|\frac{1}{h}f\left(x+\frac{h}{2}\right) - \frac{1}{h}f\left(x-\frac{h}{2}\right)-D f_0(x)\right\|_{\infty}\nonumber\\
&\leq \frac{1}{h}\left\|f_0\left(x+\frac{h}{2}\right) - \frac{h}{2}D f_0(x) - f_0\left(x-\frac{h}{2}\right)-\frac{h}{2}D f_0(x)\right\|_{\infty}\label{True part}\\
&\quad + \frac{1}{h}\left\|f\left(x+\frac{h}{2}\right)-f_0\left(x+\frac{h}{2}\right)\right\|_{\infty} + \frac{1}{h}\left\|f\left(x-\frac{h}{2}\right)-f_0\left(x-\frac{h}{2}\right)\right\|_{\infty} \label{Mixed part}
\end{align}
The term (\ref{True part}) can be bounded with the Taylor expansion of $f_0$ for some $\tau_1, \tau_2 \in [0,1]$ as follows:
\begin{align*}
    (\ref{True part}) &= \frac{1}{h}\left\|\left\{f_0(x)+\frac{h}{2}D f_0\left(x+\tau_1\frac{h}{2}\right)-\frac{h}{2}D f_0(x)\right\} - \left\{f_0(x)-\frac{h}{2}D f_0\left(x-\tau_2\frac{h}{2}\right)+\frac{h}{2}D f_0(x)\right\}\right\|_{\infty}\\
    &=\frac{1}{2}\left\|D f_0\left(x+\tau_1\frac{h}{2}\right)-D f_0(x) + D f_0\left(x-\tau_2\frac{h}{2}\right)-D f_0(x)\right\|_\infty\\
    &\leq \frac{L}{2}\left|\tau_1\frac{h}{2}\right|+\frac{L}{2}\left|\tau_2\frac{h}{2}\right| \leq \frac{L}{2}h.
\end{align*}
The expectation of the term (\ref{Mixed part}) is the $L^\infty$-risk of black-box model $f$, which is assumed to be bounded with the order of $O(\psi_n^{1/2})$. Thus, we obtain
\begin{align*}
    \mathbb{E}_{f}\left[\left\|\widehat{D} f_{0}-D f_{0}\right\|_{\infty}^2\right] &\leq
    \frac{L^2}{4}h^2 + \frac{4}{h^2}\mathbb{E}_{f}\left[\left\|f-f_0\right\|_{\infty}^2\right] + L\mathbb{E}_{f}\left[\left\|f-f_0\right\|_{\infty}\right]\\
    &\leq \frac{L^2}{4}h^2 + \frac{4}{h^2}C_0\psi_n + LC_0\psi_n^{1/2}.
\end{align*}
If we choose the bandwidth as $h_{1,n}^{*} = \alpha \psi_n^{1/4}$ for some $\alpha>0,$ we have the desired result that the squared $L^\infty$ risk of $\widehat{D} f_0$ is bounded above with the order of $O(\psi_n^{1/2})$.  
Since the convergence rate does not depend on $j$ and $\bold{x}_{-j},$ the proof is done.
$\square$

\subsection{Proof of Theorem \ref{thm:imp_est}} \label{proof:imp_est}

First, we have 
\begin{align}
    \biggl |\hat{I}&(\bold{\bold{j}})-I(\bold{\bold{j}})\biggr|\nonumber\\
    &=\biggl|\; \mathbb{E}_{\bold{X}'_{\bold{j}}\sim \hat{\mathbb{P}}_{\bold{j}}}\left[\operatorname{Var}_{\bold{X}'_{\bold{j}^c}\sim \hat{\mathbb{P}}_{\bold{j}^c}}\left\{\widehat{D}_{\bold{j}}f_{0}(\bold{X}'_{\bold{j}},\bold{X}'_{\bold{j}^c})|\bold{X}'_{\bold{j}}\right\}\right] - \mathbb{E}_{\bold{X}'_{\bold{j}}\sim \mathbb{P}_{\bold{j}}}\left[\operatorname{Var}_{\bold{X}'_{\bold{j}^c}\sim \mathbb{P}_{\bold{j}^c}}\left\{D_{\bold{j}}f_{0}(\bold{X}'_{\bold{j}},\bold{X}'_{\bold{j}^c})|\bold{X}'_{\bold{j}}\right\}\right] \;\biggr|\nonumber\\
    &\leq\biggl|\; \mathbb{E}_{\bold{X}'_{\bold{j}}\sim \hat{\mathbb{P}}_{\bold{j}}}\left[\operatorname{Var}_{\bold{X}'_{\bold{j}^c}\sim \hat{\mathbb{P}}_{\bold{j}^c}}\left\{\widehat{D}_{\bold{j}}f_{0}(\bold{X}'_{\bold{j}},\bold{X}'_{\bold{j}^c})|\bold{X}'_{\bold{j}}\right\}\right] -
    \mathbb{E}_{\bold{X}'_{\bold{j}}\sim \hat{\mathbb{P}}_{\bold{j}}}\left[\operatorname{Var}_{\bold{X}'_{\bold{j}^c}\sim \hat{\mathbb{P}}_{\bold{j}^c}}\left\{D_{\bold{j}}f_{0}(\bold{X}'_{\bold{j}},\bold{X}'_{\bold{j}^c})|\bold{X}'_{\bold{j}}\right\}\right] \;\biggr|\label{sup_bound}\\
    &\quad\quad + \biggl|\; \mathbb{E}_{\bold{X}'_{\bold{j}}\sim \hat{\mathbb{P}}_{\bold{j}}}\left[\operatorname{Var}_{\bold{X}'_{\bold{j}^c}\sim \hat{\mathbb{P}}_{\bold{j}^c}}\left\{D_{\bold{j}}f_{0}(\bold{X}'_{\bold{j}},\bold{X}'_{\bold{j}^c})|\bold{X}'_{\bold{j}}\right\}\right] -
    \mathbb{E}_{\bold{X}'_{\bold{j}}\sim \mathbb{P}_{\bold{j}}}\left[\operatorname{Var}_{\bold{X}'_{\bold{j}^c}\sim \mathbb{P}_{\bold{j}^c}}\left\{D_{\bold{j}}f_{0}(\bold{X}'_{\bold{j}},\bold{X}'_{\bold{j}^c})|\bold{X}'_{\bold{j}}\right\}\right] \;\biggr|.\label{empiric_conv}
\end{align}
The term (\ref{empiric_conv}) is about the convergence of the moments of the true derivative $D_{\bold{j}}f_0$ from the empirical distribution $\hat{\mathbb{P}}$ to the true distribution $\mathbb{P}$ which converges to 0 with the order of $n^{-1/2}$ and so we 
focus on (\ref{sup_bound}).
The term (\ref{sup_bound}) can be decomposed again and individually bounded with the term of $\|\widehat{D}_{\bold{j}}f_0-D_{\bold{j}}f_0\|_{\infty}$ as follows:
\begin{align}
    (\ref{sup_bound})&\leq\biggl|\;\mathbb{E}_{X'_{\bold{j}}\sim \hat{\mathbb{P}}_{\bold{j}}}\left[\mathbb{E}_{X'_{\bold{j}^c}\sim \hat{\mathbb{P}}_{\bold{j}^c}}\left\{
    (\widehat{D}_{\bold{j}}f_{0}(X'_{\bold{j}},X'_{\bold{j}^c}) - D_{\bold{j}}f_{0}(X'_{\bold{j}},X'_{\bold{j}^c}))^2|X'_{\bold{j}}
    \right\}\right]\;\biggr|\label{a}\\
    &\quad\quad + 2\;\biggl|\;\mathbb{E}_{X'_{\bold{j}}\sim \hat{\mathbb{P}}_{\bold{j}}}\left[ \mathbb{E}_{X'_{\bold{j}^c}\sim \hat{\mathbb{P}}_{\bold{j}^c}}\left\{
    D_{\bold{j}}f_{0}(X'_{\bold{j}},X'_{\bold{j}^c})(\widehat{D}_{\bold{j}}f_{0}(X'_{\bold{j}},X'_{\bold{j}^c}) - D_{\bold{j}}f_{0}(X'_{\bold{j}},X'_{\bold{j}^c})|X'_{\bold{j}}
    \right\}\right]\;\biggr|\label{b}\\
    &\quad\quad + \biggl|\;\mathbb{E}_{X'_{\bold{j}}\sim \hat{\mathbb{P}}_{\bold{j}}}\left[ \mathbb{E}_{X'_{\bold{j}^c}\sim \hat{\mathbb{P}}_{\bold{j}^c}}\left\{
    \widehat{D}_{\bold{j}}f_{0}(X'_{\bold{j}},X'_{\bold{j}^c}) - D_{\bold{j}}f_{0}(X'_{\bold{j}},X'_{\bold{j}^c})|X'_{\bold{j}}
    \right\}^2\right]\;\biggr|\label{c}\\
    &\quad\quad + 2\;\biggl|\;\mathbb{E}_{X'_{\bold{j}}\sim \hat{\mathbb{P}}_{\bold{j}}}\left[\mathbb{E}_{X'_{\bold{j}^c}\sim \hat{\mathbb{P}}_{\bold{j}^c}}\left\{
    D_{\bold{j}}f_{0}(X'_{\bold{j}},X'_{\bold{j}^c})|X'_{\bold{j}}\right\}\mathbb{E}_{X'_{\bold{j}^c}\sim \hat{\mathbb{P}}_{\bold{j}^c}}\left\{(\widehat{D}_{\bold{j}}f_{0}(X'_{\bold{j}},X'_{\bold{j}^c}) - D_{\bold{j}}f_{0}(X'_{\bold{j}},X'_{\bold{j}^c})|X'_{\bold{j}}
    \right\}\right]\;\biggr|\label{d}
\end{align}
and 
\begin{align*}
    (\ref{a}) &= \int\int_{\mathcal{X}_j\times\mathcal{X}_{\bold{j}^c}}(\widehat{D}_{\bold{j}}f_{0}(X'_{\bold{j}},X'_{\bold{j}^c}) - D_{\bold{j}}f_{0}(X'_{\bold{j}},X'_{\bold{j}^c}))^2d\hat{\mathbb{P}}_{\bold{j}}d\hat{\mathbb{P}}_{\bold{j}^c} = \left\|\widehat{D}_{\bold{j}}f_{0} - D_{\bold{j}}f_{0}\right\|^2_n\leq\left\|\widehat{D}_{\bold{j}}f_{0} - D_{\bold{j}}f_{0}\right\|^2_{\infty}\\
    (\ref{b}) &\leq 2L \left | \int\int_{\mathcal{X}_j\times\mathcal{X}_{\bold{j}^c}} \widehat{D}_{\bold{j}}f_{0}(X'_{\bold{j}},X'_{\bold{j}^c}) - D_{\bold{j}}f_{0}(X'_{\bold{j}},X'_{\bold{j}^c}) d\hat{\mathbb{P}}_{\bold{j}}d\hat{\mathbb{P}}_{\bold{j}^c} \right | \leq 2L \left\|\widehat{D}_{\bold{j}}f_{0} - D_{\bold{j}}f_{0}\right\|_{\infty}\\
    (\ref{c}) &\leq \int\int_{\mathcal{X}_j\times\mathcal{X}_{\bold{j}^c}}(\widehat{D}_{\bold{j}}f_{0}(X'_{\bold{j}},X'_{\bold{j}^c}) - D_{\bold{j}}f_{0}(X'_{\bold{j}},X'_{\bold{j}^c}))^2d\hat{\mathbb{P}}_{\bold{j}}d\hat{\mathbb{P}}_{\bold{j}^c}\ = \left\|\widehat{D}_{\bold{j}}f_{0} - D_{\bold{j}}f_{0}\right\|^2_n\leq\left\|\widehat{D}_{\bold{j}}f_{0} - D_{\bold{j}}f_{0}\right\|^2_{\infty}\\
    (\ref{d}) &\leq 2L \left | \int\int_{\mathcal{X}_j\times\mathcal{X}_{\bold{j}^c}} \widehat{D}_{\bold{j}}f_{0}(X'_{\bold{j}},X'_{\bold{j}^c}) - D_{\bold{j}}f_{0}(X'_{\bold{j}},X'_{\bold{j}^c}) d\hat{\mathbb{P}}_{\bold{j}}d\hat{\mathbb{P}}_{\bold{j}^c} \right | \leq 2L \left\|\widehat{D}_{\bold{j}}f_{0} - D_{\bold{j}}f_{0}\right\|_{\infty}.\\
\end{align*}
 Thus $\biggl |\hat{I}(\bold{j})-I(\bold{j})\biggr|$ can be bounded with the order of the leading term $\left\|\widehat{D}_{\bold{j}}f_{0} - D_{\bold{j}}f_{0}\right\|_{\infty}$ and so the proof is done.
 $\square$

\subsection{Extension of Theorem \ref{thm:conti}  to binary input features} \label{app: cate}
In this subsection, we prove that Theorem \ref{thm:conti} is still valid for binary input features
if we replace the partial derivative operator by the partial difference operator.
Suppose that  $\mathcal{X} =\{0,1\}^{p}.$ We assume the functional ANOVA model for $f_0$ as:
\begin{align}
f_0(\bold{x}) = \beta_{0} + \sum_{k=1}^{p}\bigg\{ \sum_{\bold{j} \in J_{k}}\beta_{\bold{j}}\bold{x}_{\bold{j}}! \bigg\},
\label{eq: cate}
\end{align}
where $\bold{x}_{\bold{j}}!=\prod_{l\in \bold{j}} x_l.$
Note that when $x_{j}$s are all binary, i.e. 0 or 1, any $f_0(\bold{x})$ can be represented by this ANOVA model. For $\bold{j} \subset [p]$, $\beta_{\bold{j}} \neq 0$ implies that the interaction $\bold{j}$ is significant.
For a given $\bold{j} \subset [p]$, we write equation (\ref{eq: cate}) as follows:
\begin{align}
f_0(\bold{x})=\beta_{0} + \sum_{\bold{j}' \subset \bold{j}}\bold{x}_{j'}! \bigg\{ \beta_{\bold{j'}} + \sum_{\bold{j}_{2} \subset \bold{j}^{c}}\beta_{\bold{j'}\cup \bold{j}_{2}}\bold{x}_{\bold{j}_{2}}! \bigg\} + \sum_{\bold{j}_{3}\subset \bold{j}^{c}}\beta_{\bold{j}_{3}}\bold{x}_{\bold{j}_{3}}!\nonumber
\end{align}
For given $\bold{j}$ and $\bold{j}'\subset \bold{j}$, let $g_{\bold{j}',\bold{j}}(\bold{x}_{\bold{j}^{c}}) = \beta_{\bold{j}'} + \sum_{\bold{j}_{2} \subset \bold{j}^{c}} \beta_{\bold{j}' \cup \bold{j}_{2} }\bold{x}_{\bold{j}_{2}}!$. Then, we can write 
\begin{align}
f_0(\bold{x}) = \beta_{0} + \sum_{\bold{j}' \subset \bold{j}} \bold{x}_{\bold{j}'}!g_{\bold{j}'\cup \bold{j}}(\bold{x}_{\bold{j}^{c}}) + \sum_{\bold{j}_{3} \subset \bold{j}^{c}}\beta_{\bold{j}_{3}}\bold{x}_{\bold{j}_{3}}!\nonumber
\end{align}
\begin{theorem} \label{theorem: category}
For a given $\bold{j}$, $\beta_{\bold{j}'}=0$ for all $\bold{j}' > \bold{j}$ if and only if $g_{\bold{j}',\bold{j}}(\bold{x}_{\bold{j}^{c}})$ is a constant function for all $\bold{x}_{\bold{j}^{c}}.$ 
\end{theorem}

({\it Proof})

The `if' part is trivial. Therefore, we only prove the `only if' part. First, it holds that $g_{\bold{j},\bold{j}}(\bold{x}_{\bold{j}^{c}})=\beta_{\bold{j}}$ when $\bold{x}_{\bold{j}^{c}}=\bold{0}.$ In turn, we can show that $\beta_{\bold{j}'} = 0$ for any $\bold{j}' > \bold{j}$ and $|\bold{j}' \backslash \bold{j}|=1$ by letting $\bold{x}_{\bold{j}^{c}}$ such that $x_{j}=1$ for $j \in \bold{j}' \backslash \bold{j}$ and 0 otherwise. By applying similar arguments repeatedly, we can show that $\beta_{\bold{j}'}=0$ for all $\bold{j}' > \bold{j}$. $\square$

For a given $\bold{j},$ we define $g(\bold{x}_{\bold{j}^c})=
f_0(\bold{x}:\bold{x}_{\bold{j}}=\bold{v})$
for any $\bold{v}\in \{0,1\}^{|\bold{j}|}$
as the function of $\bold{x}_{\bold{j}^c}$ whose output
is $f_0(\bold{x})$ with $\bold{x}_{\bold{j}}=\bold{v}.$

\begin{theorem}
For any $\bold{j}$, we can represent $g_{\bold{j}',\bold{j}}(\bold{x}_{\bold{j}^{c}})$ as follows.
\begin{align}
g_{\bold{j}',\bold{j}}(\bold{x}_{\bold{j}^{c}}) = \sum_{\bold{j}' \subseteq \bold{j}}(-1)^{|\bold{j} \backslash \bold{j}'|}f_0(\bold{x} : \bold{x}_{\bold{j}'} = \bold{1}, \bold{x}_{\bold{j} \backslash \bold{j}'} =\bold{0}).\nonumber
\end{align}
\label{th:cat2}
\end{theorem}

({\it Proof)}

Let $\bold{j} = (j_{1},...,j_{k})$ where $k=|\bold{j}|$. Since
$f_0(\bold{x} : \bold{x}_{j} = 1) = \beta_{0} + g_{\bold{j},\bold{j}} + \sum_{\bold{j}' < \bold{j}} g_{\bold{j}',\bold{j}}(\bold{x}_{\bold{j}^{c}}) + \sum_{\bold{j}_{3}\subset \bold{j}^{c}}\beta_{\bold{j}_{3}}\bold{x}_{\bold{j}_{3}}!$ and $f_0(\bold{x}:\bold{x}_{j} = 0) = \beta_{0} + \sum_{\bold{j}_{3} \subset \bold{j}^{c}}\beta_{\bold{j}_{3}}\bold{x}_{\bold{j}_{3}}!$, the following holds:
\begin{align}
g_{\bold{j},\bold{j}}(\bold{x}_{\bold{j}^{c}}) = f_0(\bold{x} : \bold{x}_{j} =1) - f_0(\bold{x}:\bold{x}_{j} = 0 ) - \sum_{\bold{j}' < \bold{j}} g_{\bold{j}',\bold{j}}(\bold{x}_{\bold{j}^{c}})\nonumber
\end{align}
Considering that
\begin{align}
f_0(\bold{x} : \bold{x}_{\bold{j}'} = \bold{1}, \bold{x}_{\bold{j} \backslash \bold{j}'} =\bold{0}) = \beta_{0} + \sum_{\Tilde{\bold{j}} \subseteq \bold{j}' }g_{\Tilde{\bold{j}},\bold{j}}(\bold{x}_{\bold{j}^{c}}) + \sum_{\bold{j}_{3} \subset \bold{j}^{c}}\beta_{\bold{j}_{3}}\bold{x}_{\bold{j}_{3}}! \label{eq: cate_first}
\end{align}
we can get
\begin{align}
\sum_{\Tilde{\bold{j}} \subseteq \bold{j}' }g_{\Tilde{\bold{j}},\bold{j}}(\bold{x}_{\bold{j}^{c}}) = f_0(\bold{x} : \bold{x}_{\bold{j}'} = \bold{1}, \bold{x}_{\bold{j} \backslash \bold{j}'} = \bold{0}) - f_0(\bold{x}: \bold{x}_{j} = 0) \label{eq: cate_eq}
\end{align}
By the principle of inclusion-exclusion, we can represent $\sum_{\bold{j}' < \bold{j}}g_{\bold{j}',\bold{j}}(\bold{x}_{\bold{j}^{c}}) $ given as:

\begin{align}
\sum_{\bold{j}' < \bold{j}}g_{\bold{j}',\bold{j}}(\bold{x}_{\bold{j}^{c}}) &= (-1)^{0}\sum_{1\leq i_{1} \leq k}\sum_{\Tilde{\bold{j}}\subset \bold{j} \backslash \{j_{i_{1}} \} }g_{\Tilde{\bold{j}},\bold{j} }(\bold{x}_{\bold{j}^{c}})\nonumber \\
&+ (-1)^{1}\sum_{1\leq i_{1} < i_{2} \leq k}\sum_{\Tilde{\bold{j}} \subset \bold{j} \backslash \{j_{i_{1}},j_{i_{2}}\}  }g_{\Tilde{\bold{j}},\bold{j}}(\bold{x}_{\bold{j}^{c}})\nonumber \\
&\:\: \vdots \nonumber\\
&+ (-1)^{k-2}\sum_{1\leq i_{1}<i_{2}<\cdots<i_{k-1} \leq k }\sum_{\Tilde{\bold{j}} \subset \bold{j} \backslash \{j_{i_{1}},...,j_{i_{k-1}} \} }g_{\Tilde{\bold{j}},\bold{j}}(\bold{x}_{\bold{j}^{c}}) \label{eq: cate_summ}
\end{align}
By (\ref{eq: cate_eq}) and (\ref{eq: cate_summ}), we have the following:
\begin{align}
\sum_{\bold{j}' < \bold{j}}g_{\bold{j}',\bold{j}}(\bold{x}_{\bold{j}^{c}}) &= (-1)^{0}\sum_{1\leq i_{1} \leq k}f_0(\bold{x}:\bold{x}_{\bold{j} \backslash \{ j_{i_{1}} \} } = \bold{1}, \bold{x}_{j_{i_{1}}} = 0) \nonumber\\
&+ (-1)^{1}\sum_{1\leq i_{1} < i_{2} \leq k}f_0(\bold{x} : \bold{x}_{\bold{j} \backslash \{ j_{i_{1}}, j_{i_{2}}\} } = \bold{1}, \bold{x}_{j_{i_{1}},j_{i_{2}}} = \bold{0} ) \nonumber\\
&\:\: \vdots \nonumber\\
&+ (-1)^{k-2} \sum_{1\leq i_{1} < \cdots i_{k-1} }f_0(\bold{x} : \bold{x}_{\bold{j} \backslash \{ j_{i_{1}},...,j_{i_{k-1}} \} } = \bold{0}) \nonumber\\
&- 
\begin{cases}
0  & \text{if}\:k\:\text{is}\:\text{odd} \\
2f_0(\bold{x} : \bold{x}_{\bold{j}} = 0 ) & \text{if}\:k\:\text{is}\:\text{even}
\end{cases} \label{eq: cate_last}
\end{align}

Thus by combining (\ref{eq: cate_first}) and (\ref{eq: cate_last}), the proof is completed. 
\begin{flushright}
$\square$
\end{flushright}

Theorem \ref{th:cat2} implies that $g_{\bold{j},\bold{j}}(\bold{x}_{\bold{j}^{c}})$ 
is the partial difference of $f_0$ with respect to $\bold{j}$ and  
Theorem \ref{theorem: category} implies that the partial difference is constant if and only if 
$\beta_{\bold{j}'}=0$ for all $\bold{j}'>\bold{j}.$

\section{Neural Interaction Model}
\label{app: NIM details}
Neural Additive Model (NAM) \cite{agarwal2020neural} is the state-of-the-art XAI model 
for GAM.
NAM models each term of GAM via a neural network and learns all of the neural networks
simultaneously by use of a gradient descent algorithm.

Neural Interaction model (NIM) is a straightforward extension of NAM that incorporates interactions. 
For a given partial functional ANOVA model
\begin{align}
\label{def:step2}
\begin{split}
\displaystyle 
f_{\mathcal{R}}(\bold{x})=\beta_{0} + &\sum_{k=1}^{K} \sum_{\bold{j} \in \mathcal{R}_k} 
f_{\bold{j}}(\bold{x}_{\bold{j}}),
\end{split}
\end{align}
NIM models each interaction term in $\mathcal{R}$ by a neural network and learns
all of the interactions using a gradient descent algorithm.

\section{Details of Meta-ANOVA for practical use}
\label{app: details}

To apply Meta-ANOVA, we have to choose two quantities: the sequence of thresholds $\{\gamma_k\}$
and bandwidths $\{h_{k,n}\}.$
For $\gamma_k,$ we set $\gamma_k= \tau \max_{\bold{j}\in \mathcal{C}_k} \hat{I}(\bold{j})$
for a pre-specified constant $\tau \in (0,1),$ where $\mathcal{C}_{k}$ is defined in the Algorithm \ref{alg: step1}.
For all experiments, we set $\tau=0.1.$
Note that the interaction screening algorithm of Meta-ANOVA is expected not to be affected much 
by the choice of $\tau$ since higher order interactions are deleted unless all of its ancestor interactions survive. See Section \ref{app: tau} of Appendix for sensitivity analysis for $\tau.$

For the bandwidth $h_{k,n},$ we let $h_{k,n}=0.1$, which is selected through trial and error.
Results of sensitivity analysis for the choice of the bandwidths are presented in Section \ref{sec:h} of Appendix.

For certain data, too many interactions could be selected, and so applying NIM only with selected interactions
would be still computationally demanding. To resolve this problem, we set the maximum sizes
for $\mathcal{C}_k.$ In our numerical studies, we set the maximum sizes $(300,100,20)$ for the second, third and fourth-order interactions, respectively.

\section{Measuring the importance of each component}
\label{app: score interaction}

As we mentioned, the importance score of Meta-ANOVA does not give information about the importance of each component.
For the importance of a given component $\bold{j},$ we propose to use the variance
of $\hat{f}_{\bold{j}}$ (i.e. 
$\operatorname{Var}_{\bold{X}_{\bold{j}} \sim \hat{\mathbb{P}}_{\bold{j}}} (\hat{f}_{\bold{j}}(\bold{X}_{\bold{j}})$)
as the measure of the importance of the component $\bold{j},$
where $\hat{f}_{\bold{j}}$ is the estimation of $f_{0,\bold{j}}$ given by NIM.

Care should be taken due to the issue of identifiability of interactions.
In general, each component in the functional ANOVA model is not identifiable.
For example, the model $f(x_1,x_2)=f_1(x_1)+f_2(x_2)$ can be rewritten as
$f(x_1,x_2)= f_1(x_1)/2 + f(x_2)/2 + f_{12}(x_1,x_2),$ where
$f_{12}(x_1,x_2)=(f_1(x_1)+f_2(x_2))/2.$
There are various conditions to ensure the identifiability of each component e.g., \cite{hooker2007ganova} and \cite{chong1993ssanova}.
Among those, we use the identifiability condition in \cite{chong1993ssanova}, which requires that
$\mathbb{E}_{X_l\sim \hat{\mathbb{P}}_l} f_{\bold{j}}(X_l, \mathbf{x}_{l^c}) =0 $
for all $\mathbf{x}_{l^c}$ and all $l\in \bold{j}.$

For a given estimation $\hat{f}_{\bold{j}},$ we can transform it to satisfy the identifiability condition.
We explain it for the second-order interaction, but higher order interactions can be treated similarly.
For simplicity, we consider the model
 \begin{align}
 \hat{f}(x_{1},x_{2}) = \hat{f}_{1}(x_{1}) + \hat{f}_{2}(x_{2}) + \hat{f}_{1,2}(x_{1},x_{2}).
 \end{align}
First, we make the interaction term $\hat{f}_{1,2}(x_{1},x_{2})$ satisfy the identifiability condition as follows:
$$\tilde{f}_{1,2}(x_1,x_2)= \hat{f}_{1,2}(x_1,x_2)-\mathbb{E}_{X_1\sim \hat{\mathbb{P}}_1} \hat{f}_{1,2}(X_1,x_2)
-\mathbb{E}_{X_2\sim \hat{\mathbb{P}}_2} \hat{f}_{1,2}(x_1,X_2)+ \mathbb{E}_{X_1\sim \hat{\mathbb{P}}_1}\mathbb{E}_{X_{2} \sim \hat{\mathbb{P}}_2} \hat{f}_{1,2}(X_1,X_2).$$
The next step is to transform $\hat{f}_1(x_1)+\mathbb{E}_{X_2\sim \hat{\mathbb{P}}_2} \hat{f}_{1,2}(x_1,X_2)(:= g_1(x_1))$ to satisfy the identifiability condition, which is easily done by letting
$\tilde{f}_{1}(x_1)=g_1(x_1)-\mathbb{E}_{X_1\sim \hat{\mathbb{P}}_1} g_1(X_1).$ 
We define $\tilde{f}_2(x_2)$ accordingly.
Finally,  we write it as:
\begin{align*}
\hat{f}(x_1,x_2)=\tilde{\beta}_0+\tilde{f}_1(x_1)+\tilde{f}_2(x_2)+\tilde{f}_{1,2}(x_1,x_2)
\end{align*}
for some real constant $\tilde{\beta}_0,$
where all of the components $\tilde{f}_1,\tilde{f}_2$, and $\tilde{f}_{1,2}$ satisfy the identifiability condition.
We measure the importance of each component after this identifiable transformation of a given functional ANOVA model.

\section{Details for Experiments}\label{app:details_exp}

\subsection{Experiments with synthetic datasets} \label{app: syn_details}

\begin{table}[H]
\fontsize{8pt}{8pt}
\selectfont
\centering
\caption[9pt]{Test suite of synthetic regression models.}
\label{table:synthetic_func}
\vskip 0.13in
\renewcommand{\arraystretch}{2.5}
\begin{tabular}{l|c}
\hline
$F_1$ & $\displaystyle \pi^{x_1 x_2} \sqrt{2x_3} - \sin^{-1}(x_4) + \log(x_3+x_5) - \frac{x_9}{x_{10}} \sqrt{\frac{x_7}{x_{8}}} - x_2 x_7$ \\ \hline
$F_2$ &  $\displaystyle \pi^{x_1 x_2} \sqrt{2|x_3|} - \sin^{-1}(0.5 x_4)  + \log(|x_3+x_5|+1) + \frac{x_9}{1+ |x_{10}|} \sqrt{\frac{|x_7|}{1+|x_{8}|}} - x_2 x_7 $\\ \hline
$F_3$ &  $\displaystyle \exp|x_1-x_2| + |x_2 x_3| - x_3^{2|x_4|} + \log(x_4^2+x_5^2+x_7^2+x_8^2) + x_9 + \frac{1}{1+x_{10}^2}$\\ \hline
$F_4$ &  $\displaystyle \exp|x_1-x_2| + |x_2 x_3| - x_3^{2|x_4|} + (x_1 x_4)^2+\log(x_4^2+x_5^2+x_7^2+x_8^2) + x_9 + \frac{1}{1+x_{10}^2}$\\ \hline
$F_5$ & $\displaystyle \frac{1}{1+x_1^2+x_2^2+x_3^2} + \sqrt{\exp(x_4+x_5)} + |x_6+x_7| + x_8 x_9 x_{10}$ \\ \hline
$F_6$ & $\displaystyle \exp(|x_1 x_2|+1) - \exp(|x_3+x_4|+1) + \cos(x_5+x_6-x_8) + \sqrt{x_8^2+x_9^2+x_{10}^2}$ \\ \hline
$F_7$ & $\displaystyle (\arctan(x_1)+\arctan(x_2))^2 + \max(x_3 x_4 + x_6, 0) - \frac{1}{1+(x_4 x_5 x_6 x_7 x_8)^2} + \left(\frac{|x_7|}{1+|x_9|}\right)^5 + \sum_{i=1}^{10} x_i$ \\ \hline
$F_8$ &  $\displaystyle x_1 x_2 + 2^{x_3+x_5+x_6} + 2^{x_3 + x_4 + x_5 +x_7} + \sin(x_7 \sin(x_8 + x_9)) + \arccos(0.9x_{10})$\\ \hline
$F_9$ & $\displaystyle \tanh(x_1 x_2 + x_3 x_4) \sqrt{|x_5|} + \exp(x_5 + x_6) + \log((x_6 x_7 x_8)^2 + 1) +x_9 x_{10} + \frac{1}{1+|x_{10}|}$ \\ \hline
$F_{10}$ & $\displaystyle \sinh(x_1 + x_2) + \arccos(\tanh(x_3 + x_5 + x_7)) + \cos(x_4 + x_5) + \sec(x_7 x_9)$ \\ \hline
\end{tabular}
\end{table}

Synthetic datasets are generated in the same way as NID \cite{tsang2017detecting} and PID \cite{liu2020detecting} do (Table \ref{table:synthetic_func}). 
We generate 30k data samples from the distribution as follows and divide them into training, validation, and test datasets, each of which consists of 10k samples.
In the case of $F_{1}$, we generate $x_{1},x_{2},x_{3},x_{6},x_{7},x_{9} \sim^{iid} \text{Uniform}(0,1)$ and $x_{4},x_{5},x_{8},x_{10} \sim^{iid} \text{Uniform}(0.6,1).$ 
For the other regression models $F_{2}$ to $F_{10}$, the input features $x_{1},x_{2},x_{3},x_{4},x_{5},x_{6},x_{7},x_{8},x_{9},x_{10}$ are generated independently from $\text{Uniform}(-1,1)$.
For a given input feature vector $\mathbf{x},$ we set $y=F_k(\mathbf{x})$ for all $k.$ (i.e., not adding noise).

We use the neural network with hidden node sizes (140, 100, 60, 20) as the baseline black-box model, which is identical to the model used in \cite{tsang2017detecting}.
All networks are trained via the Adam optimizer with the learning rate 5e-4 for all datasets. We set the batch size for training as 4096.

The hidden node sizes of the neural networks in NIM are set to be (32, 16).
The NIM is trained via the Adam optimizer with the learning rate 5e-4,
the weight decay 7.483e-9, and the batch size 4096.

\subsection{Experiments with real datasets} \label{app: NIM detalis for real}

\begin{table}[H]
\begin{center}
\begin{small}
\caption{Descriptions of real data.}
\label{Table : Dataset}
\begin{tabular}{ccccr}
\toprule
Dataset       & Size & Dimension of features & Problem \\
\midrule
\midrule
\texttt{Calhousing}     & 21k& 8  & Regression \\
\texttt{Abalone}        & 4k & 10  & Regression \\
\midrule
\texttt{German credit}  & 1k & 61 & Classification \\
\texttt{Online}         &40k & 58 & Classification\\
\texttt{Letter}         &20k & 16 & Classification \\
\bottomrule
\end{tabular}
\end{small}
\end{center}
\end{table}

\begin{table}[H]
\centering
\small
\caption{Prediction performances of various baseline black-box models.}
\label{Table : perfor-black}
\begin{tabular}{ccccr} 
\toprule
Dataset & Measure &XGB & RF & DNN \\
\midrule
\midrule
\texttt{Calhousing} & MSE $\downarrow$&\textbf{0.164} & 0.192  & 0.198  \\
\texttt{Abalone}   & MSE $\downarrow$&0.512  & 0.467  & \textbf{0.432}  \\
\midrule
\texttt{German credit} & AUROC $\uparrow$& 0.767 & 0.786 & \textbf{0.787}        \\
\texttt{Online}    & AUROC $\uparrow$&0.715 & \textbf{0.723} & 0.624 \\
\texttt{Letter}    & AUROC $\uparrow$&0.994 & \textbf{0.996} & 0.996 \\
\bottomrule
\end{tabular}
\end{table}

Table \ref{Table : Dataset} describes details of the 5 real datasets.
Details regarding the baseline black-box models are as follows. 
For the extreme gradient boosting(XGB), the number of base learners is 100, the maximum of depth is 6, and the learning rate is 0.1. For Random forest(RF), the number of base learners is 100, and all of the other parameters are set to the default ones.
For DNN,  we use the same architecture of DNN used in \cite{tsang2017detecting}: the hidden sizes (140, 100, 60, 20). Also, when learning DNNs, we set the learning rate to 1e-3 and the batch size to 1024.

Experimental details for NIM are as follows.
We train each component using a neural network with the hidden node sizes (64, 32).
Also, for \texttt{Abalone} and \texttt{Letter} datasets, the learning rate of the Adam optimizer is 1e-2, and for \texttt{Calhousing}, \texttt{German credit}, and \texttt{Online news}, the learning rate of the Adam optimizer is 1e-3. The weight decay is 7.483e-9 and the batch size is fixed as 1024.

\section{Additional experimental results} \label{Additional results}

\subsection{The performance results of Meta-ANOVA for various $K$} \label{app: order NIM}

Table \ref{Table : perfor based order} presents the prediction performances of Meta-ANOVA for various values of the maximum order of interactions $K.$ 
Note that for \texttt{German credit} and \texttt{Abalone}, none of the $4^{th}$ order interactions is selected, thus the results for $K=3$ and $K=4$ are identical. 
On the other hand, the selected $4^{th}$ order interactions are helpful for
for \texttt{Calhousing}, \texttt{Online news} and \texttt{Letter}.

\begin{table}[H]
\caption{Prediction performances of Meta-ANOVA with various values of the maximum order of interactions $K$}
\label{Table : perfor based order}
\vskip 0.15in
\begin{center}
\begin{small}
\begin{tabular}{ccccccr}
\toprule
Dataset & Problem &Max order : 2 & Max order : 3 & Max order : 4 \\
\midrule
\midrule
\texttt{Calhousing} & Regression &0.224 (0.01) & 0.200 (0.01) & \textbf{0.165} (0.01)  \\
\texttt{Abalone}   & Regression &0.435 (0.04)  & 0.427 (0.04)  & \textbf{0.427} (0.04)  \\
\midrule
\texttt{German credit}   & Classification & \textbf{0.778} (0.02) & 0.772 (0.02) & 0.772 (0.02)         \\
\texttt{Online}    & Classification & 0.714 (0.002) & 0.713 (0.002) & \textbf{0.720} (0.003) \\
\texttt{Letter}    & Classification &0.988 (0.001) & 0.994 (0.001) & \textbf{0.994} (0.001) \\
\bottomrule
\end{tabular}
\end{small}
\end{center}
\vskip -0.1in
\end{table}

\subsection{Performance results for various baseline black-box model}

Table \ref{Table : perfor for var black} presents the prediction performances of Meta-ANOVA
on \texttt{Abalone} data with various baseline black-box models and various $K.$
It is interesting to see  that the prediction performances of Meta-ANOVA are not much sensitive to the choice of the baseline black-box model, even when the prediction performances of baseline black-box models are not similar (e.g., See Table \ref{Table : perfor-black}). This could be because the screening procedure of Meta-ANOVA is robust to the choice of a baseline black-box model.

\vspace{-0.1in}
\begin{table}[H]
\caption{The prediction performances of Meta-ANOVA for various baseline black-box models and various max orders}
\label{Table : perfor for var black}
\vskip 0.15in
\begin{center}
\begin{small}
\begin{tabular}{ccccr}
\toprule
Baseline black-box model & Max order : 2 & Max order : 3 & Max order : 4 \\
\midrule
RF  & 0.435 (0.04) & 0.436 (0.03) & 0.431 (0.04) \\
XGB & 0.425 (0.04) & 0.418 (0.03)  & 0.427 (0.04) \\
DNN & 0.435 (0.04)  & 0.428 (0.04)  & 0.427 (0.04)\\
\bottomrule
\end{tabular}
\end{small}
\end{center}
\vskip -0.1in
\end{table}

\subsection{Comparison between SHAP and Meta-ANOVA in view of global and local interpretations} \label{app: shap}

In this subsection, we compare SHAP and Meta-ANOVA.
The original SHAP (\cite{lundberg2017unified}) is a local interpreter in the sense that SHAP measures the importance of each input feature for a given datum.
Global SHAP (\cite{molnar2022interpretable2nd}) is defined as the average of the absolute values of local SHAP values for all data.

We can define the local and global importance measures of each input feature based on the functional ANOVA model.
For local importance, we use $\phi_j(\mathbf{x},f)= \sum_{\mathbf{j}:j\in \mathbf{j}} f_{\mathbf{j}}(\mathbf{x}_{\mathbf{j}})$ for a given functional ANOVA model $f(\mathbf{x})=\beta_0+\sum_{\mathbf{j}} f_{\mathbf{j}}(\mathbf{x}_{\mathbf{j}}).$ 
This definition is a modification of local SHAP in the sense that it is equal to
local SHAP when $f$ is a generalized additive model (due to Corollary 1 of \cite{lundberg2017unified}). 
For global interpretation, we use the variance of $\phi(\mathbf{X},f)$ where $\mathbf{X}\sim \hat{\mathbb{P}}.$ We refer to those local and global importance measures as `ANOVA-SHAP'.

We demonstrate the comparison experiments on the \texttt{Calhousing} dataset (Table \ref{Table : des of cal}).
The SHAP values are calcuated with the XGB, and ANOVA-SHAP is calculated with Meta-ANOVA approximating XGB.

\begin{table}[H]
\centering
\caption{Description of input features in Calhousing.}
\label{Table : des of cal}
\begin{tabular}{@{}cccc@{}}
\toprule
Feature number & Feature name & Description & Feature type \\ \midrule
1 & MedInc & Median income in block & Numerical \\
2 & HouseAge & Median house age in block & Numerical \\
3 & AveRooms & Average number of rooms & Numerical \\ 
4 & AveBedrms & Average number of bedrooms & Numerical  \\ 
5 & Population & Population in block & Numerical \\
6 & AveOccup & Average house occupancy & Numerical \\ 
7 & Latitude & Latitude of house block & Numerical   \\ 
8 & Longitude & Longitude of house block & Numerical  \\ \bottomrule
\end{tabular}
\end{table}

\paragraph{Global interpretation.} 
For global importance of SHAP and Meta-ANOVA, we normalize the importances of each input feature by dividing them by the maximum importance.
Table \ref{Table : des of cal} presents the description of input features of \texttt{Calhousing} dataset. Table \ref{Table: results of SHAP and META} presents the global importances of SHAP and Meta-ANOVA of all input features of \texttt{Calhousing} dataset. 
The three most important features are the same for SHAP and Meta-ANOVA, even though the importances for the other input features are different. In practice, we can use Meta-ANOVA as a tool to confirm the validity of global SHAP.

\begin{table}[H]
\centering
\caption{Feature importance of SHAP and Meta-ANOVA}
\label{Table: results of SHAP and META}
\begin{tabular}{@{}ccccccccc@{}}
\toprule
Model (Method) $\backslash$ Feature number & 1 & 2 & 3 & 4 & 5 & 6 & 7 & 8 \\ \midrule
XGB (SHAP)        & 0.759 & 0.094 & 0.211 & 0.055 & 0.057  & 0.379 & 1.000 & 0.861 \\
Meta-ANOVA (ANOVA-SHAP)   &  0.383 & 0.003 & 0.009 & 0.000 & 1e-4  & 4e-6 & 0.962 & 1.000 \\ \bottomrule
\end{tabular}
\end{table}

\paragraph{Local interpretation.}

The SHAP values for each input feature are  computed via the ``\texttt{shap}'' package in the python. Table \ref{table:local_case} presents the results of local interpretation for the following input feature vectors. Note that `Case 1' was chosen among to point out that there are many points that the two interpretations are quite similar, and `Case 2' and `Case 3' were randomly selected. 
$$\text{Case 1} : \:\: \bold{x}=(0.068, 0.961, 0.022, 0.023, 0.043, 0.004, 0.557, 0.212)'$$ 
$$\text{Case 2} : \:\: \bold{x}=(0.266, 0.980, 0.014, 0.021, 0.015, 0.001, 0.167, 0.599)'$$
$$\text{Case 3} : \:\: \bold{x}=(0.207, 0.490, 0.039, 0.032, 0.005, 0.003, 0.678, 0.433)'$$

 

\begin{table}[H]
\centering
\scriptsize
\caption{Local interpretation comparison for three cases.}
\label{table:local_case}
\begin{tabular}{@{}cccccccccc@{}}
\toprule
                        & Model (Method) $\backslash$ Feature number  & 1 & 2 & 3 & 4 & 5 & 6 & 7 & 8 \\ \midrule
\multirow{2}{*}{Case 1} & XGB (SHAP)           & -0.506 & -0.017 & -0.061 & 0.023 & -0.002 & -0.161 & -0.491 & 0.277\\
                        & Meta-ANOVA(ANOVA-SHAP) & -0.508 & 0.096 & -0.028 & 0.000 & 1e-9 & 3e-5 & -0.912 & 0.277 \\ \midrule
\multirow{2}{*}{Case 2} & XGB (SHAP)             & 0.145 & 0.150 & -0.292 & 0.031 & 0.010 & 0.805 & 0.662 & 0.045 \\
                        & Meta-ANOVA(ANOVA-SHAP) &  0.142 & 0.163 & -0.035 & 0.000 & 0.008 & 0.004 & 0.979 & 0.571 \\ \midrule
\multirow{2}{*}{Case 3} & XGB (SHAP)             & -0.210 & -0.036 & 0.106 & -0.023 & -0.056 & 0.113 & -0.946 & 0.228 \\
                        & Meta-ANOVA(ANOVA-SHAP) & -0.035 &-0.005 & 0.054 & 0.000 & 0.017 & 0.001 & -0.200 & 0.083\\ \bottomrule
\end{tabular}
\end{table}

\subsection{Illustration of the functional relations of the main effects for \texttt{Calhousing}}

\begin{figure}[H]
    \centering
    \includegraphics[scale=0.2]{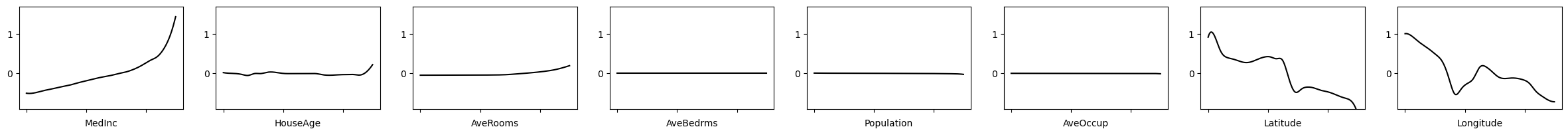}
    \centering
    \includegraphics[scale=0.2]{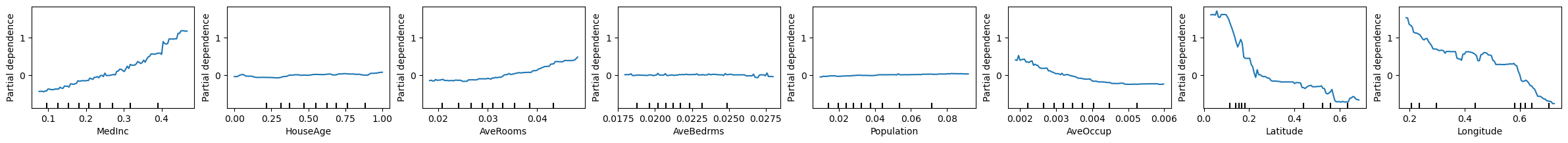}
    \caption{\texttt{Calhousing}: Partial dependence plots from XGB}
    \label{fig: calhousing shape function}
\end{figure}

In this subsection, we illustrate the functional relations of the main effects given by Meta-ANOVA
(after identifiable transformation) for 
\texttt{Calhousing} data compared to  those given by XGB.
The feature names and their descriptions are given in Table \ref{Table : des of cal}.
Figure \ref{fig: calhousing shape function} draws the functional relations of the main effects of Meta-ANOVA (black solid-line, top 8 figures) and XGB (blue solid-line, bottom 8 figures).
Most of the functional relations are similar. 
Some seemingly unnecessary bumps are observed in the functional relations of
the features `Latitude' and `Longitude' obtained by Meta-ANOVA.
These might be because NIM does not yield smooth functions. 
A post-smoothing would be required for better interpretation.

\section{Ablation studies}
\label{app: ablation}

\subsection{The number of selected interactions for various values of $\tau$} \label{app: tau}
In this subsection, we investigate the numbers of selected interactions by Meta-ANOVA with 
various values of $\tau\in \{0.1, 0.2, 0.3, 0.4,0.5\}$. 
Tables \ref{Table: num cal} to \ref{Table: num online} present the number of selected interactions for the 5 datasets. 
Recall that we limit the maximum numbers of selected interactions for each order of interactions by 300, 100, and 20 for the second, third and fourth. 
For \texttt{Calhousing}, \texttt{German credit}, and \texttt{Abalone}, 
Meta-ANOVA successively deletes unnecessary high order interactions even with $\tau=0.1.$ 
In contrast, many $4$th order interactions are selected 
in \texttt{Online news} and \texttt{Letter} even with $\tau=0.1.$
That is, the functional ANOVA model would  not be sufficient to explain \texttt{Online news} and \texttt{Letter}.
However, the numbers of selected interactions are not much affected by the choice of $\tau.$

\begin{table}[H]
\centering
\caption{\texttt{Calhousing} : The number of selected interactions for various $\tau$}
\label{Table: num cal}
\begin{tabular}{@{}cccccc@{}}
\toprule
  $\tau$     & 0.1 & 0.2 &  0.3 &  0.4 &  0.5  \\ \midrule
Main   & 8       & 8       & 8       & 8       & 8       \\ 
Second & 28      & 21      & 21      & 15      & 10      \\ 
Third  & 56      & 30      & 19      & 13      & 5       \\ 
Fourth & 20      & 10      & 6       & 0       & 1       \\ \bottomrule
\end{tabular}
\end{table}

\begin{table}[H]
\centering
\caption{\texttt{Abalone} : The number of selected interactions for various $\tau$}
\begin{tabular}{@{}cccccc@{}}
\toprule
  $\tau$     & 0.1 & 0.2 &  0.3 &  0.4 &  0.5  \\ \midrule
Main   & 10       & 10       & 10       & 10       & 10       \\ 
Second & 36      & 15      & 15      & 3      & 3      \\
Third  & 19      & 15      & 6      & 0      & 0       \\
Fourth & 0      & 0      & 0       & 0       & 0       \\ \bottomrule
\end{tabular}
\end{table}

\begin{table}[H]
\centering
\caption{\texttt{German credit} : The number of selected interactions for various $\tau$}
\begin{tabular}{@{}cccccc@{}}
\toprule
  $\tau$     & 0.1 & 0.2 &  0.3 &  0.4 &  0.5  \\ \midrule
Main   & 61       & 61       & 61       & 61       & 61       \\ 
Second & 136      & 21      & 3      & 3      & 3      \\ 
Third  & 20      & 7      & 0      & 0      & 0       \\ 
Fourth & 0      & 0      & 0       & 0       & 0       \\ \bottomrule
\end{tabular}
\end{table}

\begin{table}[H]
\centering
\caption{\texttt{Letter} : The number of selected interactions for various $\tau$}
\begin{tabular}{@{}cccccc@{}}
\toprule
  $\tau$     & 0.1 & 0.2 &  0.3 &  0.4 &  0.5  \\ \midrule
Main   & 16       & 16       & 16       & 16       & 16       \\ 
Second & 120      & 120      & 91      & 55      & 55      \\ 
Third  & 98      & 97      & 97      & 97      & 96       \\ 
Fourth & 20      & 20      & 20       & 20       & 19       \\ \bottomrule
\end{tabular}
\end{table}

\begin{table}[H]
\centering
\caption{\texttt{Online news} : The number of selected interactions for various $\tau$}
\label{Table: num online}
\begin{tabular}{@{}cccccc@{}}
\toprule
  $\tau$     & 0.1 & 0.2 &  0.3 &  0.4 &  0.5  \\ \midrule
Main   & 58       & 58       & 58       & 58       & 58       \\ 
Second & 300      & 300      & 300      & 300      & 253      \\ 
Third  & 99      & 99      & 99      & 97      & 98       \\ 
Fourth & 20      & 20      & 20       & 20       & 20       \\ \bottomrule
\end{tabular}
\end{table}

\subsection{Selection of bandwidth $h$}
\label{sec:h}

We conduct performance evaluation of Meta-ANOVA when using different bandwidths $h_{k,n}$
for $k=1$ and $k=2$ by analyzing \texttt{Abalone}.
Table \ref{table: bandwidth sensitivity} presents the MSEs for different choices of the bandwidths, which
suggests that the degree of approximation is not affected much by the choice of the bandwidth.

\begin{table}[H]
\centering
\caption{Prediction performances of Meta-ANOVA for various choices of the bandwidth $h_{k,n}.$}
\label{table: bandwidth sensitivity}
\begin{tabular}{@{}ccccc@{}}
\toprule
$h_{1,n}$ for second interaction $\backslash$ $h_{2,n}$ for main effect  & 0.05 & 0.10 & 0.15 & 0.20  \\ \midrule
0.05 & 0.3935  & 0.395  & 0.3955  & 0.3965 \\
0.10 & 0.3982  & 0.3894  & 0.3988  & 0.3916  \\
0.15 & 0.3886  & 0.3829  & 0.3839  & 0.3990  \\
0.20 & 0.3832  & 0.3877  & 0.3823  & 0.3897  \\ \bottomrule
\end{tabular}
\end{table}

\subsection{Computational complexity of interaction screening algorithm} \label{app:compute_interact_experiment}
We conduct an experiment to investigate the computational complexity of Meta-ANOVA when the dimension of input features becomes larger.
We use the synthetic regression model  $F_{6}$ for generating data. We generate 
30K input feature vectors of dimension $p$ from the uniform distribution on (-1,1). 
Then, we use the first 10 input features to generate the output. That is, only $x_{1},...,x_{10}$ are signals and the rest are non-informative.

Table \ref{table:comp_interaction} compares the computation times of interaction screening
of Meta-ANOVA for varying values of $p$ and the maximum order of interactions.
The results are normalized so that the computation time for $p=50$ and the maximum order 2 becomes 1.
It is noted that computation time does not increase exponentially as the max order increases, which confirms our conjecture that $|\mathcal{S}_k|$ decreases accordingly. 
Computation times increase as the dimension of the input feature vector increases, but it does increase linearly.  
The results amply support that Meta-ANOVA is a computationally not-heavy algorithm, applicable to large size data without much difficulty.

\begin{table}[H]
\centering
\caption{Computation times for various dimensions of input features and maximum orders.}
\label{table:comp_interaction}
\begin{tabular}{@{}cccc@{}}
\toprule
Input dimension $\backslash$ Max order & 2 & 3 & 4 \\ \midrule
50 & 1.00 & 1.12 & 1.23  \\ 
100 & 2.97 & 6.02 & 12.92  \\
150 & 6.04 & 6.12 & 6.18 \\
200 & 9.76 & 13.51 & 13.55  \\ 
250 & 15.92 & 23.33 & 23.41 \\
300 & 20.79 & 20.94 & 21.02 \\ \bottomrule
\end{tabular}
\end{table}

\subsection{Computational complexity of Meta-ANOVA}\label{app:compute_complex}

We conduct experiments to confirm how much interaction screening can save computation times.
We analyze \texttt{German credit} data by Meta-ANOVA and NIM without interaction screening.
The two hidden layer neural networks of node sizes [32, 16] is used for NIM and
is trained 300 epochs. Note that computation times of Meta-ANOVA include the times for both interaction screening and NIM learning. The device used to train is RTX 3090.

Tables \ref{table:num_comp} compares the two models in terms of the numbers of interactions and computation times (in the parenthesis), when the maximum order of interactions is set to be 2, 3, and 4, respectively. The numbers of selected components in `NIM without screening' is equal to the whole number of components including main effects and interactions, which shows how fast the number of all possible interactions increase.
Even for the maximum order 3, there are more than 30 thousands possible interactions and so
NIM is not even applicable due to computing resource limitation (marked as `-' in table). 
In contrast, Meta-ANOVA only selects about 100 interactions
even for the maximum order 4, and so can identify signal high order interactions without much difficulty.

\begin{table}[H]
\centering
\caption{The numbers of selected components (computation times)}
\label{table:num_comp}
\begin{tabular}{@{}cccc@{}}
\toprule
Model $\backslash$ Max order & 2 & 3 & 4 \\ \midrule
Meta-ANOVA & 58 (204 sec)  & 97 (306 sec) & 115 (703 sec) \\
NIM without screening & 1,891 (2,361 sec) & 37,881 (-) & 559,736 (-) \\ \bottomrule
\end{tabular}
\end{table}


\subsection{Effects of interaction screening}

One may argue that interaction screening may lose some important interactions and thus results
in accuracy loss.
In this subsection, we investigate how interaction screening affects 
the prediction performance of an approximated model.
For this purpose, we compare the performance of Meta-ANOVA both with and without interaction screening.
We construct a functional ANOVA model approximating the baseline black-box model by NIM  
with all candidate interactions being included, and compare it with the approximated model obtained by Meta-ANOVA that incorporates interaction screening.
We analyze \texttt{Abalone} and \texttt{German credit} datasets. 
Due to the large number of input features in \texttt{German credit} data, we only include the input features `account check status", `duration in month', `credit amount', and `age' in the analysis.

\begin{figure}[h]
    \centering
    \includegraphics[scale=0.5]{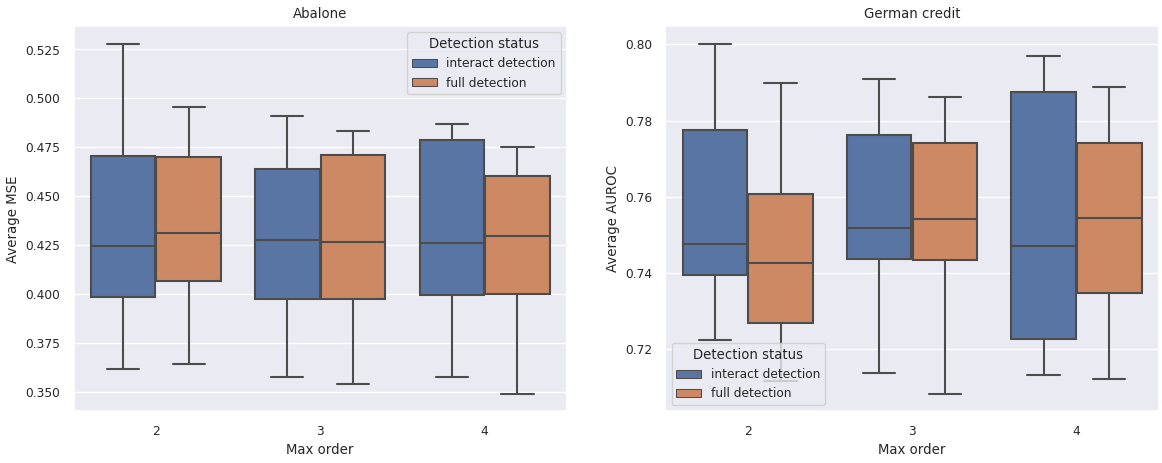}
    \vskip -0.12in
    \caption{Effects of interaction screening}
    \label{fig: Interaction_effect}
\end{figure}

Figure \ref{fig: Interaction_effect} draws the box-plots of the performance measures
of the approximated functional ANOVA models with and without interaction screening,
obtained by 10 random partitions of train/validation/test data.
It is apparent that the performances are similar even if interaction screening makes the variation of performance measures slightly larger. Moreover, for \texttt{German credit} data with $K=2,$ 
interaction screening even improves the accuracy. 
These results confirm that the interaction screening algorithm of Meta-ANOVA works quite well.

\subsection{Application of Meta-ANOVA to more complex black-box Models}
To demonstrate the applicability of Meta-ANOVA to more complex black-box models,
we consider TabTransformer (\cite{huang2020tabtransformer}) as the baseline model.
TabTransformer is an extension of the Transformer architecture that enables its application to tabular datasets.
We train a TabTransformer with depth 12 and 4 heads on \texttt{German credit} data 
and approximate it using Meta-ANOVA with the maximum order of interaction 3, whose results are presented in Table \ref{table:tap}.
As shown in Table \ref{table:tap}, Meta-ANOVA approximates the black-box model well. 

\begin{table}[H]
\centering
\caption{Prediction performances of Meta-ANOVA with  TabTransformer}
\label{table:tap}
\begin{tabular}{@{}ccc@{}}
\toprule
Model & TabTransformer(baseline) & Meat-ANOVA \\ \midrule
AUROC & 0.777 & 0.777 \\ \bottomrule
\end{tabular}
\end{table}

\subsection{Application of Meta-ANOVA to Image data} \label{app:image_data}

We apply Meta-ANOVA to image data by use of the Concept Bottleneck Model (CBM, \cite{koh2020concept}), similar to that used in \cite{radenovic2022neural}.
In CBM, instead of feeding the embedding vector obtained from image data through a CNN directly into a classifier, the CNN first predicts specific attributes or concepts associated with each image. 
These predicted attribute values are then used as inputs for the final classifier.
We consider a model for the final classifier as a black-box baseline model for Meta-ANOVA.

We analyze \texttt{CelebA} data, in which 
each image is associated with information on 40 attributes.
We use "gender" as the target label and remaining attributes as concepts related to images.
We use pretrained Resnet-18 model for CBM and DNN for the final classifier. We apply Meta-ANOVA with the maximum order of interactions 2.

Table \ref{table:perfor_celeba} presents prediction performances of CBM and Meta-ANOVA, and we can see that Meta-ANOVA approximates DNN classifier in the CBM well.
Moreover, Table \ref{table:celeba_global} presents the 5 most important attributes selected by Meta-ANOVA.
For selecting important attributes, we use the importance scores defined in Section \ref{app: score interaction} of Appendix  and normalize them by the highest score.

\begin{figure}[H]
    \centering
    \includegraphics[scale=0.2]{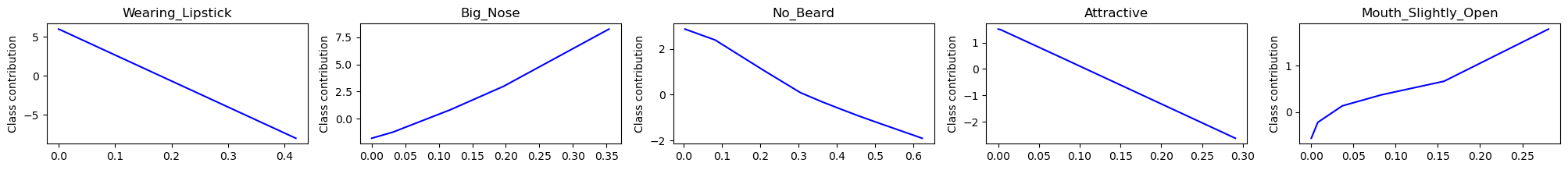}
    \caption{Functional relations of the main effects of the 5 most important attributes}
    \label{fig:plot_celeba}
\end{figure}

Figure \ref{fig:plot_celeba} describes the plots of the functional relations of the main effects
of the 5 most important attributes, all of which are quite linear.
Since all concepts are binary, Figure \ref{fig:plot_celeba} gives an interpretation for the black-box model: The more an image has a `Big Nose' and `Mouth slightly open', the more likely the image is classified as male. Conversely, if an image has 'Wearing Lipstick', 'No beard', and 'Attractive', it is more likely to be classified as female.

\begin{table}[H]
\centering
\caption{Prediction performances  of CBM and Meta-ANOVA on \texttt{CelebA} data}
\label{table:perfor_celeba}
\begin{tabular}{@{}ccc@{}}
\toprule
\textbf{Model} & CBM (baseline)  & Meta-ANOVA \\ \midrule
\textbf{Accuracy} & 0.984 & 0.983 \\ \bottomrule
\end{tabular}
\end{table}

\begin{table}[H]
\centering
\caption{Importance scores of the 5 most important attributes in \texttt{CelebA} data}
\label{table:celeba_global}
\begin{tabular}{@{}cccccc@{}}
\toprule
\textbf{Concept name} & Wearing Lipstick & Big Nose & No Beard & Attractive  & Mouth Slightly open  \\ \midrule
\textbf{Score} & 1.000 & 0.395 & 0.082 & 0.073 & 0.018   \\ \bottomrule
\end{tabular}
\end{table}

\begin{figure}[H]
    \centering
    \includegraphics[scale=0.35]{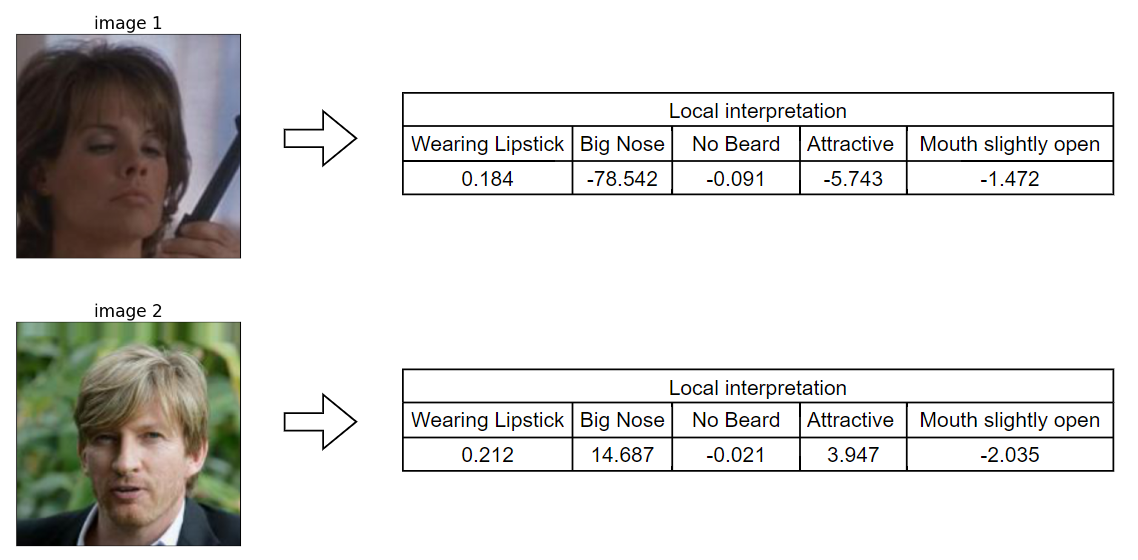}
    \caption{Results of local interpretation on CelebA data}
    \label{fig:image-local}
\end{figure}

Figure \ref{fig:image-local} compares
the local importances of the 5 attributes of two images: one is female and the other is male.
Local importances of Meta-ANOVA are defined in Section \ref{app: score interaction} of Appendix.
We can see that `Big nose' is the main reason to separate out these two images.

\
\subsection{Application of Meta-ANOVA to text data.} \label{app:text_data}

We use \texttt{GLUE-SST2} (\cite{wang2018glue}) data, which is a representative natural language benchmark dataset for “Sentimental Analysis”, and use the pre-trained SST2-DistilBERT (\cite{sanh2019distilbert}) as the baseline black-box model for Meta-ANOVA.
To obtain the importance score $I(j)$, we first randomly sample sentences containing vocabulary $j.$
Then, we calculate the outputs of SST2-DistilBERT with and without the presence of vocabulary $j$.
Note that the outputs of SST2-DistilBERT without vocabulary $j$ are obtained by masking the embedding 
corresponding to vocabulary $j$.
The variance of the output differences is used as an estimate of $I(j).$

We set the maximum order of interactions in the screening algorithm as 2. 
Once we select interactions, we construct an approximated model
by first converting each sentence to a binary vector based on the presence of
the selected main effects and second-order interactions in the sentence and then
applying the linear logistic regression with the converted binary vector as input.
Table \ref{table:perfor_bert} describes the performance results of DistilBERT and Meta-ANOVA. 
The approximated model is inferior to the baseline black-box model, DistilBERT, in terms of prediction accuracy. However, it is not surprising since the approximation model is much simpler than DistilBERT.
Instead, we may say that DistilBERT can be approximated by a simple linear logistic regression without much degradation of
accuracy.

Table \ref{table:global_bert} presents the 5 most important vocabularies identified by Meta-ANOVA and
their importance scores (normalized by the highest score).
The results suggest that
DistilBERT considers negative words more importantly in decision-making than positive words. 
Among the second-order interactions, the interaction of `not' and `bad' is the most important interaction. 
Note that `not' and `bad' are negative words marginally but when they are present together,
they give a positive meaning.

\begin{table}[H]
\centering
\caption{Prediction performances  of DistilBERT and Meta-ANOVA}
\label{table:perfor_bert}
\begin{tabular}{@{}ccc@{}}
\toprule
\textbf{Model} & DistilBERT & Meta-ANOVA  \\ \midrule
\textbf{AUROC} & 0.971 & 0.914  \\
\textbf{Selected interactions} & - & 11,753 (main) + 107,416 (2nd-interaction)  \\ \bottomrule
\end{tabular}
\end{table}

\begin{table}[H]
\centering
\caption{The 5 most important vocabularies and their importance scores}
\label{table:global_bert}
\begin{tabular}{@{}cccccc@{}}
\toprule
\textbf{Vocabulary} & not & bad & no & less & worst \\ \midrule
\textbf{Score} & 1.000 & 0.666 & 0.466 & 0.444 & 0.378  \\ \bottomrule
\end{tabular}
\end{table}

\end{document}